\documentclass[sigconf]{acmart}
\usepackage[utf8]{inputenc}
\usepackage{subfigure}
\usepackage{multirow}
\usepackage{color, verbatim,caption}
\usepackage{hyperref}
\usepackage{enumitem}
\usepackage{dsfont}
\usepackage{algorithm}
\usepackage{algorithmicx}
\usepackage{algpseudocode}
\usepackage{amsmath}

\usepackage{array}
\newcolumntype{L}[1]{>{\raggedright\let\newline\\\arraybackslash\hspace{0pt}}m{#1}}
\newcolumntype{C}[1]{>{\centering\let\newline\\\arraybackslash\hspace{0pt}}m{#1}}
\newcolumntype{R}[1]{>{\raggedleft\let\newline\\\arraybackslash\hspace{0pt}}m{#1}}

\begin{document}


\title{Weakly-supervised Relation Extraction by \\ Pattern-enhanced Embedding Learning}

\author{Meng Qu$^1$, Xiang Ren$^2$, Yu Zhang$^1$, Jiawei Han$^1$}

\affiliation{%
  \institution{{$^1$}{University of Illinois at Urbana-Champaign, IL, USA}}
}
\affiliation{%
  \institution{{$^2$}{University of Southern California, CA, USA}}
}
\affiliation{%
  \institution{{$^1$}{\{mengqu2, yuz9, hanj\}@illinois.edu} $\quad$ {$^2$}{xiangren@usc.edu}}
}

\begin{abstract} 

Extracting relations from text corpora is an important task in text mining.
It becomes particularly challenging when focusing on weakly-supervised relation extraction, that is, utilizing a few relation instances (i.e., a pair of entities and their relation) as seeds to extract more instances from corpora.
Existing distributional approaches leverage the corpus-level co-occurrence statistics of entities to predict their relations, and require large number of labeled instances to learn effective relation classifiers. 
Alternatively, pattern-based approaches perform bootstrapping or apply neural networks to model the local contexts, but still rely on large number of labeled instances to build reliable models. 
In this paper, we study integrating the distributional and pattern-based methods in a weakly-supervised setting, such that the two types of methods can provide complementary supervision for each other to build an effective, unified model.
We propose a novel co-training framework with a distributional module and a pattern module. During training, the distributional module helps the pattern module \emph{discriminate} between the informative patterns and other patterns, and the pattern module \emph{generates} some highly-confident instances to improve the distributional module.
The whole framework can be effectively optimized by iterating between improving the pattern module and updating the distributional module. We conduct experiments on two tasks: knowledge base completion with text corpora and corpus-level relation extraction. Experimental results prove the effectiveness of our framework in the weakly-supervised setting.


\end{abstract}


\maketitle

\section{INTRODUCTION}
\label{sec::intro}

\begin{figure}
	\centering
	\includegraphics[width=0.48\textwidth]{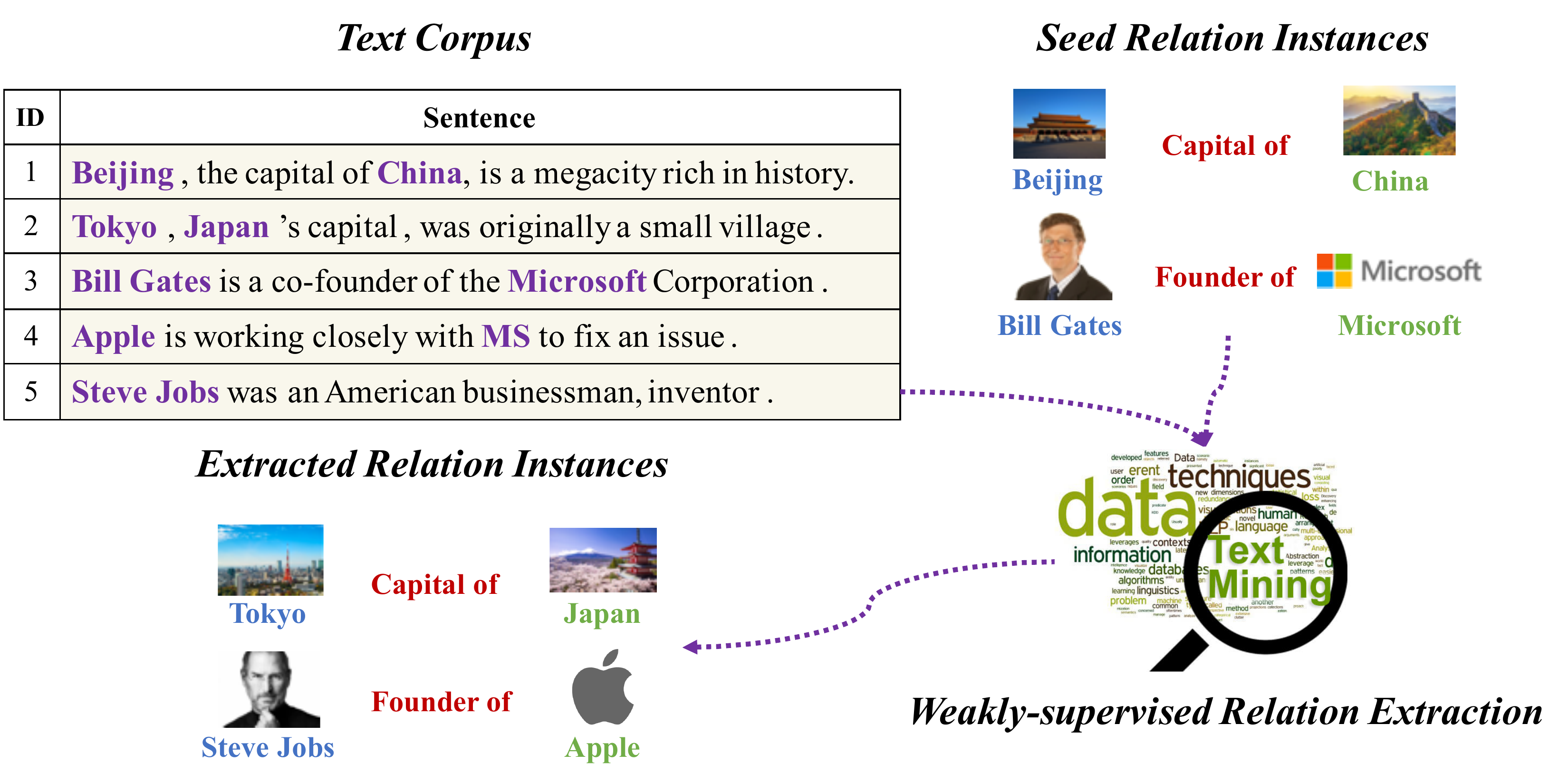}
	\caption{Illustration of weakly-supervised relation extraction. Given a text corpus and a few relation  instances as seeds, the goal is to extract more instances from the corpus.}
	\label{fig::example}
\end{figure}

Relation extraction is an important task in data mining and natural language processing. Given a text corpus, relation extraction aims at extracting a set of relation instances (i.e., a pair of entities and their relation) based on some given examples.
Many efforts~\cite{culotta2004dependency,mooney2006subsequence,ren2017cotype} have been done on sentence-level relation extraction, where the goal is to predict the relation for a pair of entities mentioned in a sentence (e.g., predict the relation between ``\textit{Beijing}'' and ``\textit{China}'' in sentence 1 of Fig.~\ref{fig::example}).
Despite its wide applications, these studies usually require a large number of human-annotated sentences as training data, which are expensive to obtain. In many cases (e.g., knowledge base completion~\cite{xu2014rc}), it is also desirable to extract a set of relation instances by consolidating evidences from multiple sentences in corpora, which cannot be directly achieved by these studies.
Instead of looking at individual sentences, corpus-level relation extraction~\cite{mintz2009distant,hoffmann2011knowledge,riedel2013relation,zeng2015distant,bing2015improving} identifies relation instances from text corpora using evidences from multiple sentences. This also makes it possible to apply weakly-supervised methods based on corpus-level statistics~\cite{agichtein2000snowball,curran2007minimising}. Such weakly-supervised approaches usually take a few relation instances as seeds, and extract more instances by consolidating redundant information collected from large corpora. The extracted instances can serve as extra knowledge in various downstream applications, including knowledge base completion~\cite{riedel2013relation,toutanova2015representing}, corpus-level relation extraction~\cite{zeng2015distant,lin2016neural}, hypernym discovery~\cite{shwartz2016improving,snow2005learning} and synonym discovery~\cite{wang2016solving,qu2017automatic}. 

In this paper, we focus on corpus-level relation extraction in the \textit{weakly-supervised setting}. There are broadly two types of weakly-supervised approaches for corpus-level relation extraction.
Among them, pattern-based approaches predict the relation of an entity pair from multiple sentences mentioning both entities. To do that, traditional approaches~\cite{nakashole2012patty,schmitz2012open,yahya2014renoun} extract textual patterns (e.g.,  tokens between a pair of entities) and new relation instances in a bootstrapping manner. However, many relations could be expressed in a variety of ways. Due to such diversity, these approaches often have difficulty matching the learned patterns to unseen contexts, leading to the problem of semantic drift~\cite{curran2007minimising} and inferior performance. For example, with the given instance ``\textit{(Beijing, Capital of, China)}'' in Fig.~\ref{fig::example}, ``\textit{[Head] , the capital of [Tail]}'' will be extracted as a textual pattern from sentence 1. But we have difficulty in matching the pattern to sentence 2 even though both sentences refer to the same relation ``\textit{Capital of}''. 
Recent approaches~\cite{xu2015classifying,liu2015dependency} try to overcome the sparsity issue of textual patterns by encoding textual patterns with neural networks, so that pattern matching can be replaced by similarity measurement between vector representations. However, these approaches typically rely on large amount of labeled instances to train effective models~\cite{shwartz2016improving}, making it hard to deal with the weakly-supervised setting. 

Alternatively, distributional approaches resort to the corpus-level co-occurrence statistics of entities. The basic idea is to learn low-dimensional representations of entities to preserve such statistics, so that entities with similar semantic meanings tend to have similar representations. With entity representations, a relation classifier can be learned using the labeled relation instances, which takes entity representations as features and predicts the relation of a pair of entities. To learn entity representations, some approaches~\cite{mikolov2013efficient,pennington2014glove,tang2015line} only consider the given text corpus. Despite the unsupervised property, their performance is usually limited due to the lack of supervision~\cite{xu2014rc}. To learn more effective representations for relation extraction, some other approaches~\cite{xu2014rc,wang2014knowledge} jointly learn entity representations and relation classifiers using the labeled instances. However, similar to pattern-based approaches, distributional approaches also require considerable amount of relation instances to achieve good performance~\cite{xu2014rc}, which are usually hard to obtain in the weakly-supervised setting. 

The pattern-based and the distributional approaches extract relations from different perspectives, which are naturally complementary to each other.
Ideally, we would wish to integrate both approaches, so that they can mutually enhance and reduce the reliance on the given relation instances. Towards integrating both approaches, several existing studies~\cite{shwartz2016improving,toutanova2015representing,qu2017automatic} try to jointly train a distributional model and a pattern model using the labeled instances. However, the supervision of their frameworks still totally comes from the given relation instances, which is insufficient in the weakly-supervised setting. Therefore, their performance is yet far from satisfaction, and we are seeking an approach that is more robust to the scarcity of seed instances.

\begin{figure}
	\centering
	\includegraphics[width=0.48\textwidth]{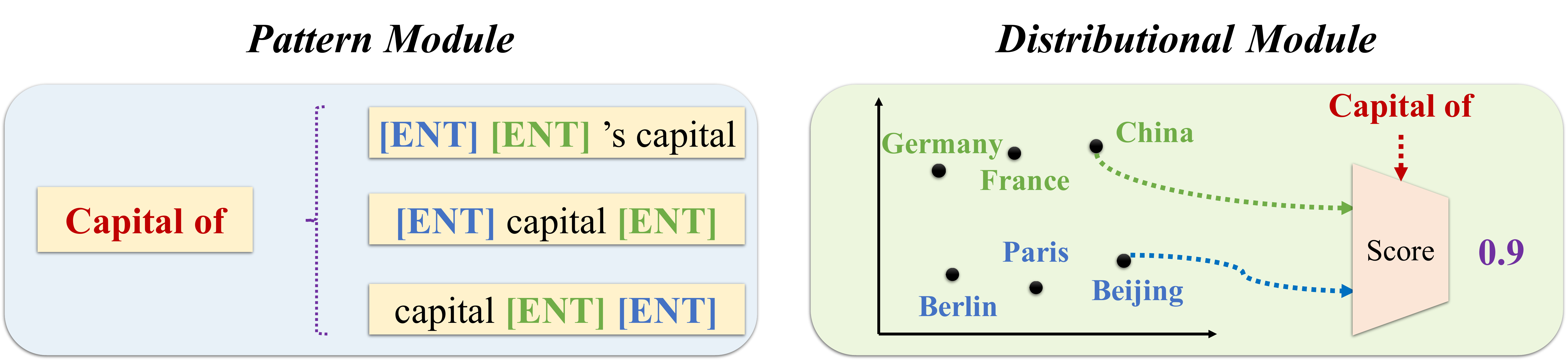}
	\caption{Illustration of the modules. The pattern module aims to learn reliable textual patterns for each relation. The distributional module tries to learn entity representations and a score function to estimate the quality of each instance.}
	\label{fig::module}
\end{figure}

\begin{figure}
	\centering
	\includegraphics[width=0.48\textwidth]{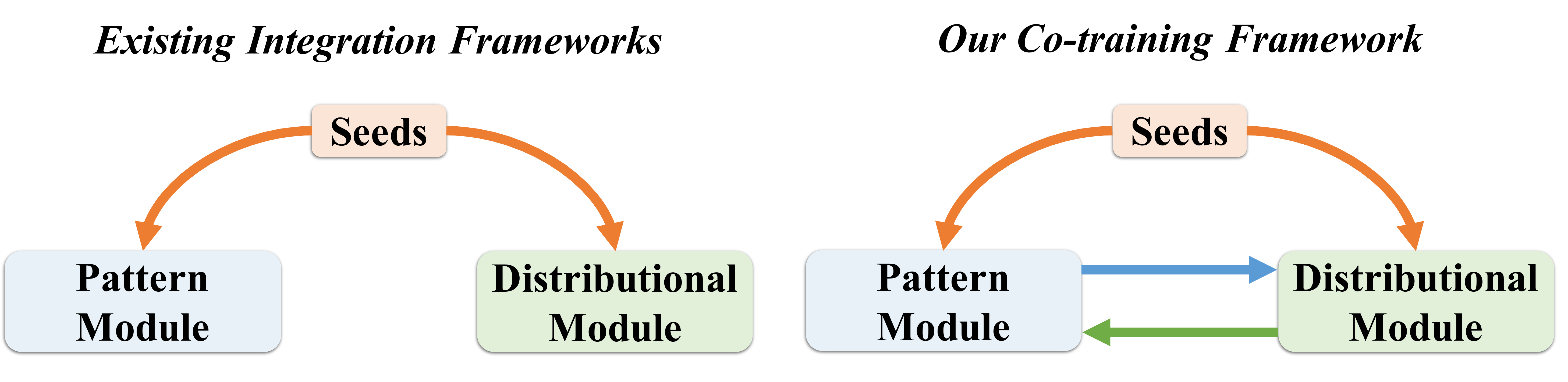}
	\caption{Comparison with existing integration frameworks. Existing frameworks totally rely on the seed instances to provide supervision. Our framework encourages both modules to provide extra supervision for each other.}
	\label{fig::comparison}
\end{figure}

In this paper, we propose such an approach called REPEL (Relation Extraction with Pattern-enhanced Embedding Learning) for weakly-supervised relation extraction. Our approach consists of a pattern module and a distributional module (see Fig.~\ref{fig::module}). 
The pattern module aims at learning a set of reliable textual patterns for relation extraction; while the distributional module tries to learn a relation classifier on entity representations for prediction. 
Different from existing studies, we follow the co-training~\cite{blum1998combining} strategy and encourage both modules to provide extra supervision for each other, which is expected to complement the limited supervision from the given seed instances (see Fig.~\ref{fig::comparison}).
Specifically, the pattern module acts as a \emph{generator}, as it can extract some candidate instances based on the discovered reliable patterns; whereas the distributional module is treated as a \emph{discriminator} to evaluate the quality of each generated instance, that is, whether an instance is reasonable. To encourage the collaboration of both modules, we formulate a joint optimization process, in which we iterate between two sub-processes. 
In the first sub-process, the discriminator (distributional module) will evaluate the instances generated by the generator (pattern module), and the results serve as extra signals to adjust the generator. In the second sub-process, the generator (pattern module) will in turn generate a set of highly confident instances, which serve as extra training seeds to improve the discriminator (distributional module). During training, we keep iterating between the two sub-processes, so that both modules can be consistently improved. Once the training converges, both modules can be applied to relation extraction, which extract new relation instances from different perspectives.


In summary, in this paper we make the following contributions:
\begin{itemize}[leftmargin=*,noitemsep,nolistsep]
	\item We propose a principled framework to integrate the distributional and pattern-based methods for weakly-supervised relation extraction, which is effective in overcoming the scarcity of seeds.
	\item We develop a joint optimization algorithm for solving the unified objective, alternating between adjusting the pattern module and improving the distributional module.
	\item We conduct experiments on two downstream applications over two real-world datasets. Experimental results prove the effectiveness of our framework in the weakly-supervised setting.
\end{itemize}

\section{PROBLEM DEFINITION}
\label{sec::definition}

In this section, we formally define our problem.

\smallskip
\noindent \textsf{\textbf{Entity Name.}}
An \emph{entity name} is a string referring to a real-world entity, which usually appears in multiple sentences of a corpus. For example in Fig.~\ref{fig::example}, all strings with purple colors (e.g., Beijing, Bill Gates) are valid entity names.

To extract relations between different entities, a prerequisite is to detect those entity names in text corpora. In this paper, for simplicity, we will not focus on entity name detection. Instead, we will use existing tools to do that. Specifically, we first apply some named entity recognition tools~\cite{manning-EtAl:2014:P14-5} to the corpus, which are able to detect entity names in text. In practice, many detected entity names can refer to the same entity. For example in Fig.~\ref{fig::example}, ``\textit{Microsoft}'' in sentence 3 and ``\textit{MS}'' in sentence 4 both refer to \textit{Microsoft Corporation}. For entity names representing the same entity, since they have exactly the same meaning, we may expect to treat them equally instead of treating them independently. Therefore, we further leverage some entity linking tools~\cite{isem2013daiber}, which can link synonymous entity names to the same entity in an external knowledge (e.g., Freebase). After entity linking, for each entity, we use a unified id to replace all entity names referring to that entity. For example, we can use the Freebase id of the entity \textit{Microsoft Corporation} to replace ``\textit{Microsoft}'' and ``\textit{MS}'' in Fig.~\ref{fig::example}.

\smallskip
\noindent \textsf{\textbf{Relation Instance.}}
A \emph{relation instance} describes the relation between a pair of entities. Formally, a relation instance is composed of an entity pair $(e_h, e_t)$ and a relation $r$, meaning that entity $e_h$ and entity $e_t$ have the relation $r$.

Relation instances are ubiquitous. For example in Fig.~\ref{fig::example} \emph{(Beijing, China)} with \emph{capital of}, \emph{(Bill Gates, Microsoft)} with \emph{founder of} are both valid relation instances. Extracting such instances from text corpora is an essential task, which has wide applications.

\smallskip
\noindent \textsf{\textbf{Problem Definition.}}
In this paper, we study weakly-supervised relation extraction. Specifically, given a text corpus $D$ and some target relations $R$, with each target relation $r$ specified by a set of relation instances $\{(e_h^{r(k)}, e_t^{r(k)}, r)\}_{k=1}^{N_r}$, our goal is to leverage the given instances as seeds and extract more instances from the corpus (Fig.~\ref{fig::example}). Formally, we define our problem as follows:

\begin{definition}
\label{def::problem}
\textbf{(Problem Definition)}
\textsl{\textbf{Given} a text corpus $D$ and some target relations $R$, where each target relation $r$ is characterized by a few seed instances $\{(e_h^{r(k)}, e_t^{r(k)}, r)\}_{k=1}^{N_r}$ or in other words a few seed entity pairs $\{(e_h^{r(k)}, e_t^{r(k)})\}_{k=1}^{N_r}$, the weakly-supervised relation extraction task \textbf{aims to} extract more instances $\{(e_h^{(i)}, e_t^{(i)}, r^{(i)})\}_{i=1}^{M}$ from the corpus. In other words, we aim at discovering more entity pairs $\{(e_h^{r(i)}, e_t^{r(i)})\}_{i=1}^{M_r}$ under each target relation $r \in R$.}
\end{definition}

\section{THE REPEL FRAMEWORK}
\label{sec::model}

\subsection{Framework Overview}

In this section, we introduce our approach to weakly-supervised relation extraction. The major challenge comes from the deficiency of supervision, since we only have a few relation instances as seeds. Therefore, the performances of existing approaches, including the pattern-based~\cite{agichtein2000snowball,zeng2014relation,lin2016neural} and the distributional approaches~\cite{mikolov2013distributed,bordes2013translating,xu2014rc}, are not satisfactory. Although some studies~\cite{shwartz2016improving,toutanova2015representing,qu2017automatic} trying to reduce the reliance on seeds by integrating both approaches, they simply employ a joint training framework, which still requires considerable relation instances to train effective models.

To better overcome the challenge of seed scarcity, in this paper we propose a framework called REPEL based on the co-training strategy~\cite{blum1998combining}. Our framework consists of two modules, a pattern module and a distributional module (see Fig.~\ref{fig::module}), which extract relations from different perspectives. The pattern module aims at finding a set of reliable textual patterns for relation extraction. Meanwhile, the distributional module tries to learn entity representations and train a score function, which measures the quality of a relation instance. Different from existing studies, both modules are encouraged to provide extra supervision to each other, which is expected to complement the limited supervision from seed instances (see Fig.~\ref{fig::comparison}). Specifically, the pattern module is treated as a generator since it can extract some candidate relation instances, and meanwhile the distributional module acts as a discriminator to evaluate each instance. During training, the discriminator evaluates the instances generated by the generator, and the results serve as extra signals to adjust the generator. On the other hand, the generator will in turn generate some highly confident instances, which act as extra seeds to improve the discriminator. We keep iterating between adjusting the pattern module and improving the distributional module. Once the training process converges, both modules can be utilized to discover more instances.

The overall objective is summarized below:
\begin{equation}
\label{eqn::objective}
	\max_{P,D} O = \max_{P,D} \{ O_p + O_d + \lambda O_i \}.
\end{equation}
In the objective, $P$ represents the parameters of the pattern module, that is, a given number of reliable patterns for each target relation. $D$ denotes the parameters of the distributional module, that is, entity representations and a score function.
The objective function consists of three terms. Among them, $O_p$ is the objective of the pattern module, in which we leverage the given seed instances for pattern selection. $O_d$ is the objective of the distributional module, which learns relevant parameters under the guidance of seed instances. Finally, $O_i$ models the interactions of both modules.


Next, we introduce the model details. \textbf{Note that} for simplicity, we only consider one relation when introducing the model. To deal with multiple relations, we can simply combine their objectives.


\subsection{Pattern Module}

\begin{figure}
	\centering
	\includegraphics[width=0.48\textwidth]{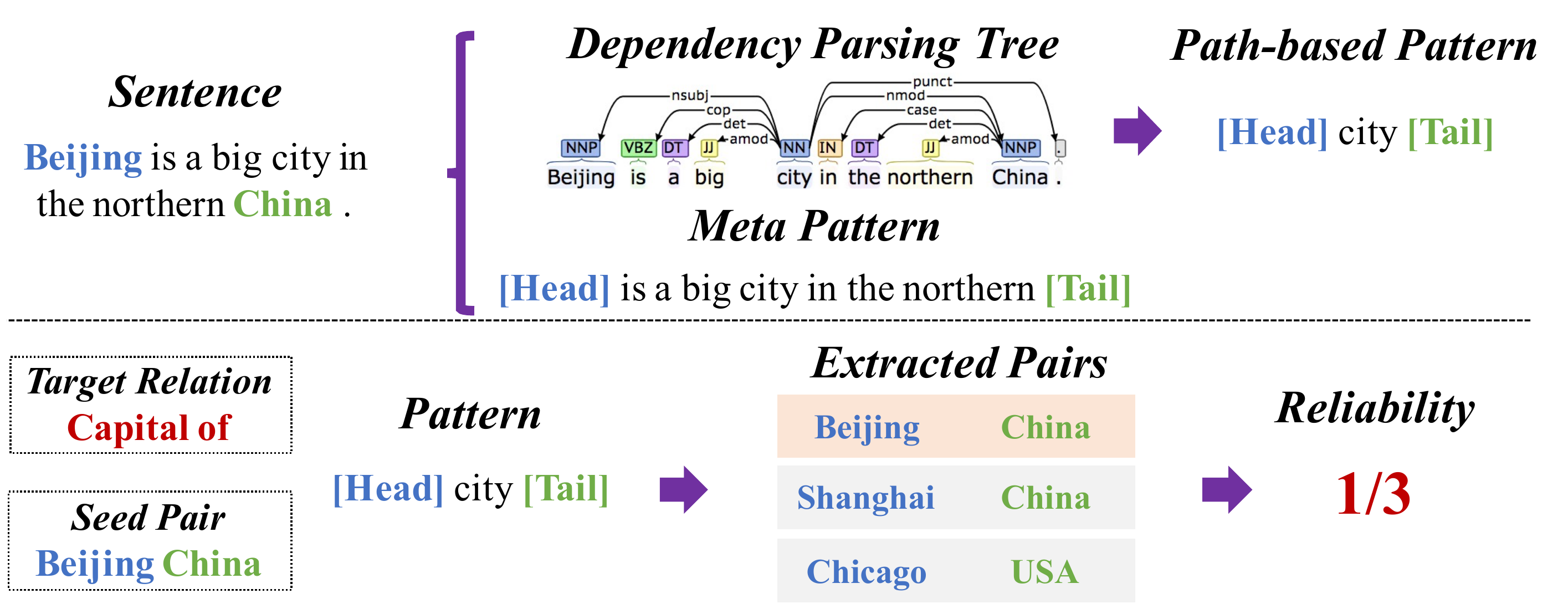}
	\caption{Illustration of the pattern module. We consider both the path-based pattern and meta pattern. We infer pattern reliability using the seed entity pairs.}
	\label{fig::pattern}
\end{figure}

In the pattern module, our goal is to select a given number of the most reliable patterns $P$ for the target relation, and further leverage them to discover more relation instances from corpora.


Following previous studies on pattern-based approaches, we leverage both the path-based patterns~\cite{bunescu2005shortest,nakashole2012patty,xu2015classifying} and the meta patterns~\cite{jiang2017metapad}. For a pair of entities in a sentence, the path-based pattern is defined as the tokens along the shortest dependency path between the two entities. Whereas the meta pattern is defined as a sequence of context words around the entities. Fig.~\ref{fig::pattern} presents an example of both patterns. Given the definition of patterns, we can go back to the corpus and extract patterns for every pair of entities in a sentence, forming a set of candidate patterns and many entity pairs linked to each pattern.

Among all the candidate patterns, we hope to extract the most reliable ones for the target relation. Towards this goal, we leverage the seed relation instances as guidance, and estimate the reliability of a pattern $\pi$ with the following measurement $R(\pi)$:
\begin{equation}
\label{eqn::reliability}
	R(\pi) = \frac{|G(\pi) \cap S_{pair}|}{|G(\pi)|},
\end{equation}
where $G(\pi)$ represents all the entity pairs extracted by the pattern $\pi$, and $S_{pair}$ is the set of seed entity pairs under the target relation. The numerator of $R(\pi)$ is the number of seed entity pairs which can be discovered by the pattern $\pi$, and the denominator counts all extracted entity pairs. 
For example in the right part of Fig.~\ref{fig::pattern}, we focus on the relation \emph{capital of}, and the pattern \emph{[Head] city [Tail]} extracts three entity pairs. Among them, the red pair is in the seed set, and therefore the reliability is 1/3. Such definition of $R(\pi)$ is quite intuitive. Basically, if a pattern can extract many seed entity pairs under the target relation, then it will be considered reliable.

Based on the measurement, we try to select the top $K$ reliable patterns, in which $K$ is a given number. Such goal can be achieved by optimizing the following objective function with respect to $P$:
\begin{equation}
\label{eqn::o_p}
	O_p = \sum_{\pi \in P} R(\pi),
\end{equation}
where $P$ is the pattern set with size $K$.

Once the most reliable patterns $P$ are learned for the target relation, we can leverage them to extract new entity pairs under the target relation. Formally, we denote the set of entity pairs extracted by the pattern set $P$ as $G(P)$, which is calculated as follows:
\begin{equation}
\label{eqn::extraction}
	G(P) = \cup_{\pi \in P} G(\pi) ,
\end{equation}
where $G(\pi)$ is the set of entity pairs extracted by pattern $\pi$.

\subsection{Distributional Module}
The distributional module of our approach focuses on the global distributional information of entities. Specifically, it aims at learning distributed entity representations from corpora, so that similar entities are likely to have similar representations. Meanwhile, we utilize the given relation instances as seeds to train a score function, which takes entity representations as features to estimate whether a relation instance is reasonable.


To learn entity representations from text corpora, we follow~\cite{tang2015pte} and build a bipartite network between all the entities and words. The weight between an entity and a word is defined as the number of sentences in which they co-occur. Then for an entity $e$ and a word $w$, we infer the conditional probability $P(w|e)$ as follows:
\begin{equation}
	\label{eqn::softmax}
	P(w|e)=\frac{\exp(\mathbf{x}_e \cdot \mathbf{c}_w)}{Z},
\end{equation}
where $\mathbf{x}_e$ is the vector representation of entity $e$, $\mathbf{c}_w$ is the embedding vector of word $w$ and $Z$ is a normalization term.

Given the estimated conditional probability $p(\cdot|e)$, we try to minimize its KL divergence from the empirical distribution $p'(\cdot|e)$ for every entity $e$, so that the distributional information can be preserved into the learned entity representations. Specifically, the empirical distribution is defined as $p'(w|e) \propto n_{w,e}$, where $n_{w,e}$ is the weight of the edge between word $w$ and entity $e$. After some simplification, we obtain the following objective function:
\begin{equation}
	\label{eqn::text}
	O_{text}=\sum_{w,e}n_{w,e}\log P(w|e),
\end{equation}
The above objective function can be efficiently optimized with the negative sampling~\cite{mikolov2013distributed} and edge sampling~\cite{tang2015line} techniques. In each epoch, a positive edge and several negative edges are sampled for optimization. For details, readers may refer to~\cite{tang2015line,tang2015pte}.

Meanwhile, we also leverage the given seed instances to learn a score function, which estimates the quality of a instance, that is, how likely an entity pair has the target relation. Following the previous work~\cite{bordes2013translating}, for an entity pair $f=(e_h,e_t)$, its score under the target relation is defined as follows:
\begin{equation}
	\label{eqn::score}
	L_D(f|r)=1-|| x_{e_h} + y_r - x_{e_t} ||_2^2,
\end{equation}
where $||\cdot||_2$ is the Euclidean norm of a vector, $x_e$ is the representation of entity $e$, $r$ is the target relation and $y_r$ is a parameter vector for the target relation.

Intuitively, we expect a seed entity pair could have larger scores than some randomly sampled pairs under the target relation. Therefore, we adopt the following ranking based objective for training:
\begin{equation}
	\label{eqn::seed}
	O_{seed} = \sum_{f \in S_{pair}} \sum_{f'=(e'_h,e'_t)}\min\{1,L_D(f|r) - L_D(f'|r)\}.
\end{equation}
$S_{pair}$ is all seed pairs, $e'_h$ and $e'_t$ are randomly sampled entities. 

Finally, we integrate Eqn.~\ref{eqn::text} and Eqn.~\ref{eqn::seed} as the objective of the distributional module, and we try to optimize it with respect to $D$.
\begin{equation}
	\label{eqn::o_d}
	O_{d} = O_{text} + \eta O_{seed},
\end{equation}
where $\eta$ is used to control the weights of the two parts, $D$ represents all parameters of the distributional module, including entity representations $x_e$ and the parameter vector $y_r$ of the relation.

Once the representations are learned, we can use the score function $L_D$ to measure the score of each entity pair under the target relation, and thus discover some highly confident relation instances.

\subsection{Modeling the Module Interaction}
So far, the supervision of both modules totally comes from the given relation instances, which is insufficient in the weakly-supervised setting. To solve this problem, we follow the co-training strategy~\cite{blum1998combining}, and encourage both modules to provide extra supervision for each other.

Specifically, we introduce the following objective function, and try to maximize it with respect to both of $P$ and $D$:
\begin{equation}
	\label{eqn::o_i}
	O_i = E_{f \in G(P)}[L_D(f|r)],
\end{equation}
where $f \in G(P)$ is an entity pair extracted by the reliable pattern set $P$ with $G(P)$ defined in Eqn.~\ref{eqn::extraction}, $L_D(f|r)$ is the score of pair $f$ under the target relation. From the objective function, we see that the selected patterns $P$ acts as a generator, since it generates some candidate entity pairs under the target relation; whereas the distributional module serves as a discriminator, trying to score the generated entity pairs under the target relation. The goal of the objective function is to encourage the agreement of the pattern module and the distributional module. More specifically, we hope that the entity pairs generated by the pattern module can be considered reasonable by the distributional module. The intuition behind the objective comes from the co-training algorithms~\cite{blum1998combining}, where it has been proved that the error rate of two predictive models can be decreased by minimizing their disagreement~\cite{chapelle2009semi}.

To intuitively understand how this objective function will improve both modules, let us consider how to optimize with respect to both modules. For the pattern module, to maximize the above objective, the pattern set $P$ should include patterns which are considered reliable by the distributional module. That is, the entity pairs generated by those patterns should obtain large scores from the distributional score function $L_D$. In this way, the distributional module provides extra supervision to estimate the pattern reliability. Meanwhile, for the distributional module, to maximize the objective function, it should assign larger scores to the entity pairs generated by the pattern module. Therefore, the highly confident entity pairs generated by the pattern module serve as extra seeds to help improve the distributional module.

With the above objective function, both modules can tightly interact with each other, and provide extra supervision to overcome the challenge of seed scarcity.

\section{THE JOINT OPTIMIZATION PROBLEM}
\label{sec::optimization}

To optimize the overall objective function (Eqn.~\ref{eqn::objective}), we leverage the coordinate gradient descent algorithm~\cite{wright2015coordinate}, by iterating between two sub-processes. In the first sub-process, we fix the pattern module, and update the distributional module under the guidance of the given seeds and the highly confident instances generated by the pattern module. In the second sub-process, the distributional module is fixed, and we update the selected patterns with the given seed instances and the supervision provided by the distributional module. During training, we keep iterating between the two sub-processes, so that both modules can be consistently improved.

\smallskip
\noindent \textsf{\textbf{1. Optimizing the Distributional Module.}}
In this step, we fix the selected pattern set $P$ to update the parameters $D$ of the distributional module.
Formally, maximizing the objective function with respect to $D$ can be transformed as the following problem:
\begin{small}
\begin{equation}
	\label{eqn::optimize_d}
	\max_{D} \{ O_d+\lambda O_i\}= \max_{D}\{ O_d + \lambda E_{f \in G(P)}[L_D(f|r)]\},
\end{equation}
\end{small}
which is a continuous optimization problem. We use the stochastic gradient descent algorithm for optimization. On the one hand, we adjust all parameters $D$ to maximize the $O_d$ part. On the other hand, some entity pairs $f$ will be sampled based on the selected patterns $P$, which are treated as extra instances to update $D$.

\smallskip
\noindent \textsf{\textbf{2. Optimizing the Pattern Module.}}
In this this, we fix the parameters $D$ of the distributional module and adjust the reliable pattern set $P$.
Formally, maximizing the objective function with respect to $P$ is equivalent to the following optimization problem:
\begin{small}
\begin{equation}
	\label{eqn::optimize_p}
	\max_{P} \{ O_p+\lambda O_i\}=\max_{P}\{ \sum_{\pi \in P} \left(R(\pi) + \lambda E_{f \in G(\pi)}[L_D(f|r)] \right) \},
\end{equation}
\end{small}
which is a discrete optimization problem, with the goal as selecting a given number of patterns $P$ with the largest reliability. The reliability of a pattern $\pi$ is calculated from two sources: $O_p$ and $O_i$. In the $O_p$ part, the reliability is measured with $R(\pi)$ defined in Eqn.~\ref{eqn::reliability}, which leverages the given seeds for reliability estimation. In the $O_i$ part, we utilize the score function $L_D$ to score each entity pair $f$ extracted by pattern $\pi$, and further average them to obtain another reliability estimation $E_{f \in G(\pi)}[L_D(f|r)]$. Finally, the two estimations are weighted as the overall reliability. In practice, we can first calculate the overall reliability of each pattern, and then select the top $K$ patterns to form the reliable pattern set $P$. 

Finally, we summarize the optimization algorithm into Alg.~\ref{algo::optimization}. Once the training converges, our approach will return a set of discovered reliable patterns from the pattern module and a distributional score function from the distributional module. Both the learned patterns and score function can be leveraged for relation extraction, which extract new instances from different perspectives. Specifically, the learned reliable patterns extract relations from local contexts by matching the contexts with the patterns, which usually have high precision but low recall. This is because for a pair of entities, the local contexts mentioning both entities are usually more reliable for predicting their relations, leading to high precision. However, for many pairs of entities, they may never co-occur in any local contexts, and thus using local contexts can result in low recall. In practice, the learned reliable patterns can be applied to applications such as corpus-level relation extraction (see the details in Sec.~\ref{sec::eval}.3 (2)). On the other hand, the learned distributional score function predict entity relation from corpus-level statistics, leading to relatively low precision but high recall, and is more suitable for tasks like knowledge base completion with text corpora (see the details in Sec.~\ref{sec::eval}.3 (1)).

\begin{algorithm}
    \caption{Optimization algorithm of REPEL.}
    \label{algo::optimization}
    \begin{algorithmic}[1]
        \Require \small{A text corpus, a few seed relation instances, the number of reliable patterns $K$, the parameter $\lambda$, the parameter $\eta$.}
        \Ensure \small{A set of reliable patterns $P$ from pattern module, a score function $L_D$ from distributional module, extracted relation instances.}
        \State \small{Generate patterns and entity pairs extracted by each pattern.}
        \State \small{Build the bipartite network between entities and words.}
        \While{not converge}
            \State \small{$\boxdot$ \textbf{\emph{Update the distributional module:}}}
            \State \small{Extract some instances by using the set of reliable patterns $P$.}
            \State \small{Optimize $D$ with both the seeds and extracted instances (Eqn.~\ref{eqn::optimize_d}).}
            \State \small{$\boxdot$ \textbf{\emph{Update the pattern module:}}}
            \State \small{Calculate pattern reliability with the seeds and $L_D$ (Eqn.~\ref{eqn::optimize_p}).}
            \State \small{Select the top $K$ most reliable patterns to form the pattern set $P$.}
        \EndWhile
        \State \small{$\boxdot$ \textbf{\emph{Extract relation instances:}}}
        \State \small{Utilize the reliable patterns $P$ to extract instances from local contexts.}
        \State \small{Utilize the distributional score function $L_D$ to extract instances.}
    \end{algorithmic}
\end{algorithm}

\section{EXPERIMENT}

In this section, we evaluate our approach on two downstream applications: knowledge base completion with text corpora (KBC) and corpus-level relation extraction (RE).

In knowledge base completion with text corpora, the key task is to predict the missing relationships between each pair of entities in knowledge bases. Since some pairs of entities may not co-occur in any sentences in the given corpus, the learned pattern module can not provide information for predicting their relations. Therefore, for KBC we only use the entity representations and score function learned by the distributional module for extraction, and we expect to show that the pattern module can provide extra seeds during training, yielding a more effective distributional module.
For corpus-level RE, it aims at predicting the relation of a pair of entities from several sentences mentioning both of them. In this case, the reliable patterns learned by the pattern module can capture the local context information from the sentences.
Therefore, we focus on utilizing the learned pattern module for prediction in RE, and we expect to show that the distributional module can enhance the pattern module by providing extra supervision to select reliable patterns.

\subsection{Experiment Setup}

\smallskip
\noindent \textsf{\textbf{1. Datasets.}}
In experiment, we leverage existing NER tool~\cite{manning-EtAl:2014:P14-5} for entity detection. Since the NER tool can only detect entities of several major types such as location,  person and organization, we thus sample 10 common relations~\footnote{\emph{location.country.capital}, \emph{people.person.parents}, \emph{people.person.children}, \emph{location.administrative division.country}, \emph{people.person.place of birth}, \emph{location.neighborhood.neighborhood of}, \emph{people.person.nationality}, \emph{people.deceased person.place of death}, \emph{location.location.contains}, \emph{organization.organization.founders}.} 
related to person, location and organization from Freebase~\footnote{~\url{https://developers.google.com/freebase/}} as our target relations.
Then two datasets are constructed based on the selected relations. \\

\begin{table}[!htb]
\caption{Statistics of the Datasets.}
	\label{tab::dataset}
	\centering
	\scalebox{0.85}{
        \begin{tabular}{c c c}
        \hline
        \textbf{Dataset} & \textbf{Wiki + Freebase} & \textbf{NYT + Freebase} \\
        \hline
        $\#$ Documents & 150,000 & 118,664  \\ 
        $\#$ Entities & 92,443 & 23,120  \\ 
        $\#$ Candidate Patterns & 621,782 & 232,892 \\ 
        $\#$ Seed Instances per Relation & 50 & 50  \\ 
        $\#$ Relations in KBC & 10 & 10  \\ 
        $\#$ Test Instances in KBC & 10,734 & 6,094  \\ 
        $\#$ Relations in RE & 5 & 6  \\ 
        $\#$ Test Entity Pairs in RE & 131 & 222  \\ 
        \hline
        \end{tabular}
    }
\end{table}

(1) \textbf{Wiki}: The first 150K articles in Wikipedia~\footnote{~\url{https://www.wikipedia.org/}} are used as the corpus. For each target relation, we randomly sample 50 relation instances from Freebase as seeds. In the knowledge base completion task, we select all the above 10 relations as the target relations, and we sample 10,734 extra instances from Freebase for prediction. In the corpus-level relation extraction task, the manually annotated sentences from~\cite{ellis2012linguistic} are used for evaluation. Among all relations in the annotated sentences, 5 relations~\footnote{\emph{people.person.children},\emph{people.person.place of birth},\emph{people.person.nationality},\emph{people.deceased person.place of death},\emph{organization.organization.founders}.} can be mapped to our selected 10 Freebase relations, and thus we only focus on these 5 relations. There are totally 194 manually annotated sentences and 131 entity pairs related to the relations. 

(2) \textbf{NYT}: The 118,664 documents from 2013 New York Times news articles. Similar to the Wiki dataset, for each target relation we randomly sample 50 relation instances from Freebase as seeds. In the knowledge base completion task, we select all the above 10 relations as the target relations, and totally 6,094 extra instances are sampled for evaluation. In the corpus-level relation extraction task, we leverage the manually annotated sentences from~\cite{etzioni2011open} for evaluation. Among all relations in the sentences, 6 relations~\footnote{\emph{people.person.children},\emph{people.person.nationality},\emph{location.location.contains},\\ \emph{people.deceased person.place of death},\emph{organization.organization.founders},\\ \emph{location.administrative division.country}.} can be mapped to the selected 10 Freebase relations, so we focus on these 6 relations. There are totally 322 manually annotated sentences and 222 entity pairs related to the relations.

For each text corpus, we adopt Stanford CoreNLP package~\cite{manning-EtAl:2014:P14-5}\footnote{~\url{http://stanfordnlp.github.io/CoreNLP/}} to do preprocessing. Then we leverage DBpedia Spotlight~\cite{isem2013daiber}\footnote{~\url{https://github.com/dbpedia-spotlight/dbpedia-spotlight}} to link the detected entity names to the Freebase.

\smallskip
\noindent \textsf{\textbf{2. Compared Algorithms.}}
In the knowledge base completion task, we select the following baseline algorithms to compare:\\
(1) \textbf{word2vec~\cite{mikolov2013distributed}}: A distributional approach for word embedding learning, which can learn entity representations from text corpora. Once the representations are learned, we utilize the seed instances to train a relation classifier (Eqn.~\ref{eqn::score}) for extraction. 
(2) \textbf{TransE~\cite{bordes2013translating}}: A distributional approach for knowledge base completion, which only uses the given seed instances for training. 
(3) \textbf{RK~\cite{wang2016solving}}: A distributional approach for knowledge base completion, which leverages both the text corpus and the given relation instances to learn entity representations. 
(4) \textbf{DPE~\cite{qu2017automatic}}: An approach that integrates the distributional and pattern-based methods. It jointly models the distributional information in text corpora, the given relation instances and the textual patterns. 
(5) \textbf{CONV~\cite{toutanova2015representing}}: A knowledge base completion approach, which integrates the distributional and pattern-based methods by jointly optimizing the given seed instances and the instances extracted by textual patterns.

In the corpus-level relation extraction task, the following approaches are selected to compare:\\
(1) \textbf{SnowBall~\cite{agichtein2000snowball}}: A pattern approach for relation extraction, which discovers reliable patterns with the seed instances in a bootstrapping way. 
(2) \textbf{PATTY~\cite{nakashole2012patty}}: A pattern approach which can apply to relation extraction. We leverage the seed instances to select relevant patterns in a bootstrapping way~\cite{agichtein2000snowball}.
(3) \textbf{CNN-ATT~\cite{lin2016neural}}: A pattern approach for corpus-level relation extraction. It leverages convolutional neural networks to encode and classify each sentence, and then consolidates the results of different sentences using an attention mechanism.
(4) \textbf{PCNN-ATT~\cite{lin2016neural}}: A pattern approach for corpus-level relation extraction. Compared with CNN-ATT, it also introduces the position embedding for each word and entity.
(5) \textbf{PathCNN~\cite{zeng2017incorporating}}: A pattern approach for corpus-level relation extraction. For each entity pair, besides sentences mentioning both entities, it also considers some other sentences mentioning only one of them.
(6) \textbf{LexNET~\cite{shwartz2016improving,shwartz2016path}}: An approach combining the distributional and pattern-based methods for relation extraction. Formally, it uses a recurrent layer to encode local textual patterns, and then uses the encoding vector together with entity representations for prediction.

For our proposed approach, we consider the following variants:\\
(1) \textbf{REPEL-P}: A variant of our approach with only the pattern module ($O_p$).
(2) \textbf{REPEL-D}: A variant of our approach with only the distributional module ($O_d$).
(3) \textbf{REPEL}: Our proposed approach, which encourages the collaboration of both modules during training. Once the training converges, we leverage the entity representations and score function learned by the distributional module for the KBC task; whereas the reliable patterns discovered by the pattern module are used for the RE task.

\label{sec::eval}
\smallskip
\noindent \textsf{\textbf{3. Evaluation Setup.}}
(1) \textbf{Knowledge Base Completion:} For each compared algorithm, we first learn entity representations and relation classifiers (or score function for our approach) by using the given text corpus and relation instances. Then the learned representations and classifiers are leveraged for evaluation. Specifically, for each test instance $(e_h, e_t, r)$, we remove its head entity or tail entity, obtaining two incomplete instances, including $(e_h, ?, r)$ and $(?, e_t, r)$, and our goal is to select the correct entity from the entity set to fill the incomplete instances. To do that, for each candidate entity in the entity set, we calculate its score by measuring the quality of the formed instance using the relation classifier. Then we sort different entities in the descending order based on their scores, and calculate the rank of the correct entity. Finally, we report the mean value of those ranks (i.e., MR) and also the proportion of the correct entities ranked within top 10 (i.e., Hits@10).\\
(2) \textbf{Corpus-level Relation Extraction:} For each compared algorithm, we first use it to predict the relation expressed in each test sentence. Specifically, for neural network based approaches (PathCNN, CNN-ATT, PCNN-ATT, LexNET), the test sentences can be directly classified based on the learned neural classifiers. For approaches based on textual patterns (PATTY, Snowball, REPEL, REPEL-P), we first match the local context of the test sentence to a discovered reliable pattern $\pi^*$, then we classify the sentence based on the relation expressed by pattern $\pi^*$. To do such matching, we represent each learned reliable pattern and the local patterns of the test sentences with a low-dimensional vector. The pattern vector is calculated by averaging the embeddings of tokens in each pattern, with the token embeddings learned by our approach in Eqn.~\ref{eqn::softmax}. Once the pattern vectors are obtained, each local pattern in test sentences is matched to its most similar reliable pattern, in which the similarity is measured as the cosine similarity between the pattern vectors.
After all test sentences are classified, for each test entity pair, we consolidate the prediction results from the test sentences mentioning both entities, and return the predicted relation together with the confidence score. During consolidate, we either average the prediction results of all test sentences (LexNET, PathCNN, PATTY, Snowball, REPEL, REPEL-P), or leverage the learned attention mechanism (CNN-ATT, PCNN-ATT). Finally, we sort all test entity pairs in the descending order based on the calculated confidence scores, and compare the ranked list with the ground-truth. Based on the results, we report both the precision at position K (i.e., P@K), recall at position K (i.e., R@K), f1 score at position K (i.e., F1@K) and the precision-recall curve.

\smallskip
\noindent \textsf{\textbf{4. Parameter Settings.}}
For all knowledge base completion methods and the distributional module of our approach, we set the dimension of all representations as 100. The number of iterations for TransE, word2vec, RK, DPE are set as 1000, 20, 20, 3B respectively to ensure the convergence. Other parameters are set as the default values suggested in the original papers.
For the neural based approaches to corpus-level relation extraction, the dimension of the embedding layer and the hidden layer is set as 100. Other parameters are set as the default values suggested in the original papers.
For our proposed approach, the parameter $\lambda$ for controlling the weight of the interaction term is set as 1 by default. For the distributional module, the learning rate is set as 0.01, the parameter $\eta$ is set as 0.005, the number of training edges in each iteration is set as 3B. For the pattern module, we set the number of reliable patterns $K$ for each relation as 100.

\subsection{Performance Comparison}

\begin{figure}[htb!]
	\centering
	\subfigure[Wiki]{
		\label{fig::hit-wiki}
		\includegraphics[width=0.22\textwidth]{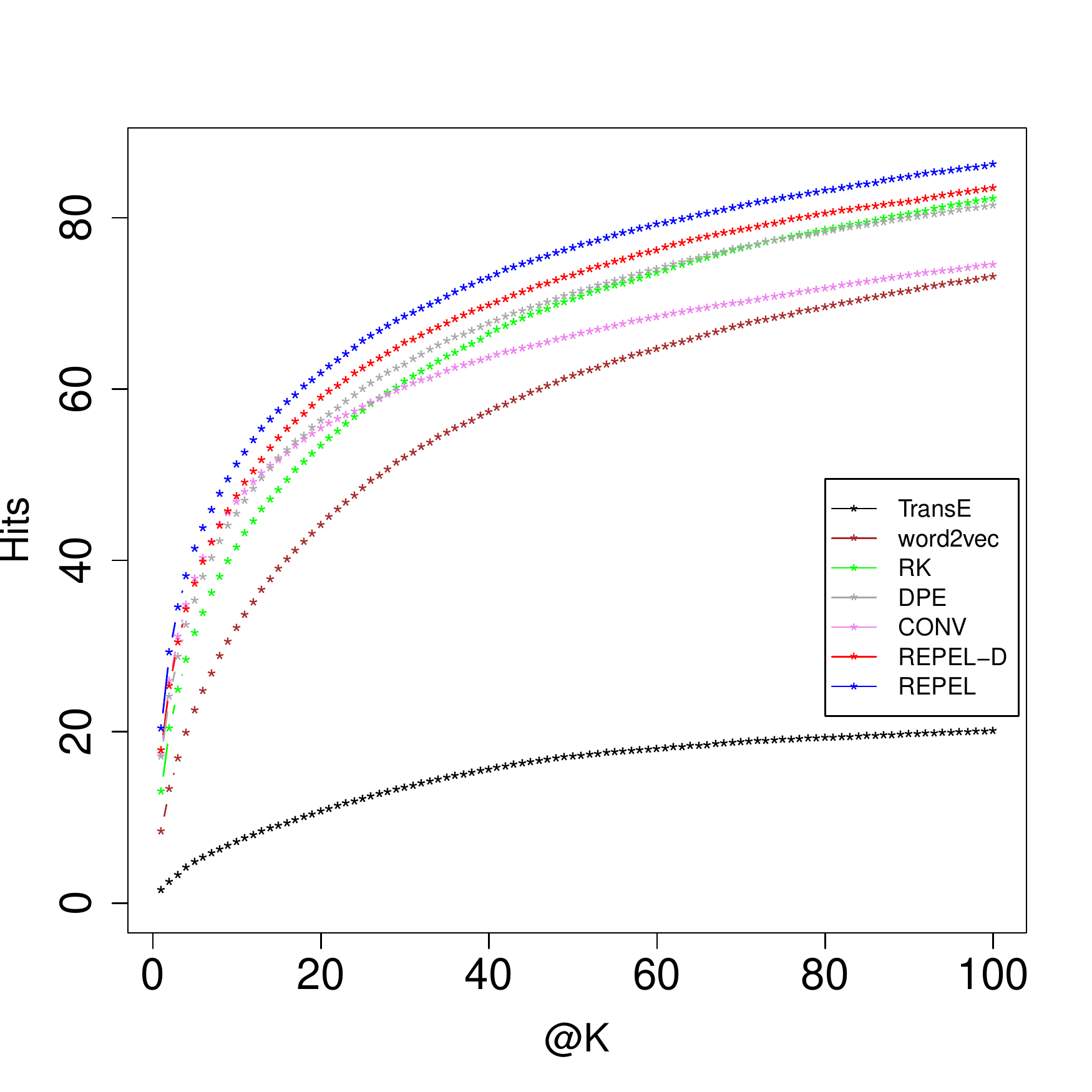}	
	}
	\subfigure[NYT]{
		\label{fig::hit-nyt}
		\includegraphics[width=0.22\textwidth]{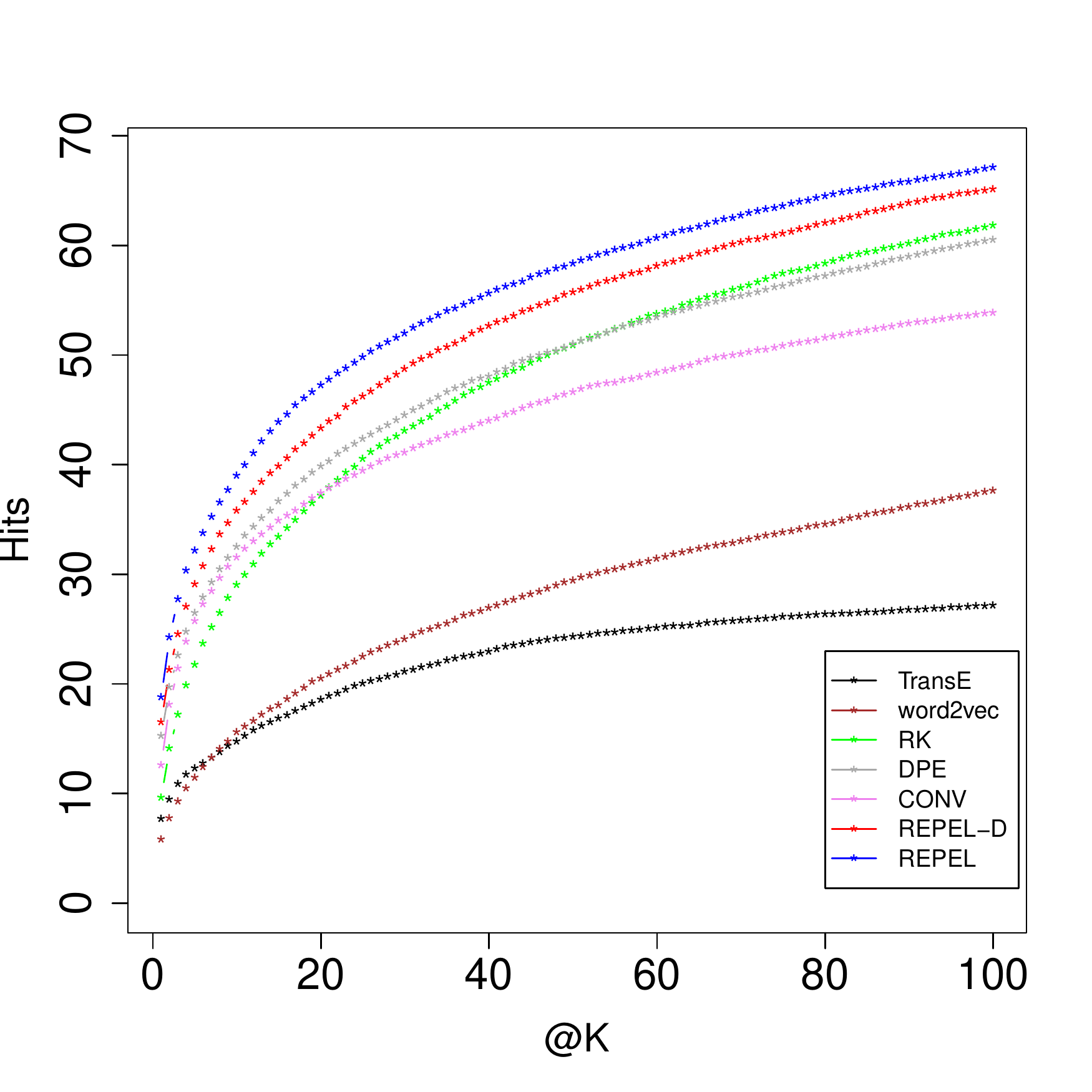}
	}
	\caption{Hits curves on the KBC task.}
	\label{fig::hit}
\end{figure}

\smallskip
\noindent \textsf{\textbf{1. Knowledge Base Completion with Text Corpora (KBC).}} 
We present the quantitative results in Table~\ref{tab::results-kbc}, and the hits curve in Fig.~\ref{fig::hit}. For the approach only considering the given seed instances (TransE), we see the performance is very limited due to the scarcity of seeds. Along the other line, the approach considering text corpora (word2vec) achieves relatively better results, but are still far from satisfactory, since it ignores the supervision from the seed instances. If we consider both the text corpus and seed instances for entity representation learning (RK), we obtain much better results. Moreover, by further jointly training a pattern model (DPE, CONV), the hits ratio can be further significantly improved.

\begin{table} [!htb]
	\caption{Quantitative results on the KBC task. }
	\label{tab::results-kbc}
	\begin{center}
		\scalebox{0.9}{
		\begin{tabular}{|C{2.3cm}|C{1.3cm}|C{1.3cm}|C{1.3cm}|C{1.3cm}|}\hline
		    \multirow{2}{*}{\textbf{Algorithm}}	& \multicolumn{2}{c|}{\textbf{Wiki + Freebase}} & \multicolumn{2}{c|}{\textbf{NYT + Freebase}} \\ \cline{2-5}
		    & \textbf{Hits@10} & \textbf{MR}& \textbf{Hits@10}& \textbf{MR} \\ \hline \hline
	        TransE~\cite{bordes2013translating}  & 7.13 & 4328.40 & 15.94 & 3833.47   \\  
            word2vec~\cite{mikolov2013distributed} & 32.12 & 203.53 & 15.56 & 913.04   \\  
	        RK~\cite{wang2016solving} & 41.49 & 72.87 & 29.01 & 307.89   \\ 
	        DPE~\cite{qu2017automatic} & 45.45 & 78.87 & 32.47 & 279.99  \\ 
    	    CONV~\cite{toutanova2015representing} & 46.84 & 139.81 & 31.51 & 903.38  \\ \hline
    	    REPEL-D & 47.49 & 67.28 & 35.79 & 234.23  \\ 
    	    REPEL & \textbf{51.18} & \textbf{62.18} & \textbf{38.98} & \textbf{199.44}  \\ \hline
	    \end{tabular}
	}
	\end{center}
\end{table}

For our proposed approach, with only the distributional module (REPEL-D), it already outperforms all the baseline approaches. Compared with DPE, the performance gain of REPEL-D mainly comes from the usage of the score function in Eqn.~\ref{eqn::score}, which can better model different relations. Compared with CONV, REPEL-D achieves better results, as the distributional information in text corpora can be better captured with Eqn.~\ref{eqn::text}. Moreover, by encouraging the collaboration of both modules (REPEL), the results are further significantly improved. This observation demonstrates that the pattern module can indeed help improve the distributional module by providing some highly confident instances.

Overall, our approach achieves quite impressive results on the knowledge base completion task compared with several strong baseline approaches. Also, the pattern module can indeed enhance the distributional module with our co-training framework.

\begin{table*} [!htb]
	\caption{Quantitative results on the RE task.}
	\label{tab::results-re}
	\begin{center}
		\scalebox{0.7}{
		\begin{tabular}{|C{2.1cm}|C{1.2cm}|C{1.2cm}|C{1.2cm}|C{1.2cm}|C{1.2cm}|C{1.2cm}|C{1.2cm}|C{1.2cm}|C{1.2cm}|C{1.2cm}|C{1.2cm}|C{1.2cm}|}\hline
		    \multirow{2}{*}{\textbf{Algorithm}}	& \multicolumn{6}{c|}{\textbf{Wiki + Freebase}} & \multicolumn{6}{c|}{\textbf{NYT + Freebase}} \\ \cline{2-13}
		    & \textbf{\small{P@50}} &\textbf{\small{R@50}} & \textbf{\small{F1@50}} & \textbf{\small{P@100}} &\textbf{\small{R@100}} & \textbf{\small{F1@100}} & \textbf{\small{P@50}} & \textbf{\small{R@50}}& \textbf{\small{F1@50}} & \textbf{\small{P@100}} &\textbf{\small{R@100}} & \textbf{\small{F1@100}}\\ \hline \hline
	        Snowball~\cite{agichtein2000snowball} & 58.00 & 22.14 & 32.05 & 65.00 & 49.62 & 56.28 & 20.00 & 4.50 & 7.35 & 21.00 & 9.46 & 13.04  \\  
	        PATTY~\cite{nakashole2012patty} & 60.00 & 22.90 & 33.15 & 61.00 & 46.56 & 52.81 & 28.00 & 6.31 & 10.30 & 20.00 & 9.01 & 12.42  \\
            CNN-ATT~\cite{lin2016neural} & 26.00 & 9.92 & 14.36 & 22.00 & 16.79 & 19.05 & 24.00 & 5.41 & 8.83 & 29.00 & 13.06 & 18.01 \\  
	        PCNN-ATT~\cite{lin2016neural} & 58.00 & 22.14 & 32.05 & 36.00 & 27.48 & 31.17 & 46.00 & 10.36 & 16.91 & 26.00 & 11.71 & 16.15   \\ 
	        PathCNN~\cite{zeng2017incorporating} & 36.00 & 13.74 & 19.89 & 38.00 & 29.01 & 32.90 & 42.00 & 9.46 & 15.44 & 26.00 & 11.71 & 16.15  \\ 
	        LexNET~\cite{shwartz2016improving,shwartz2016path} & 74.00 & 28.24 & 40.88 & 61.00 & 46.56 & 52.81 & 32.00 & 7.21 & 11.77 & 26.00 & 11.71 & 16.15 \\  \hline
	        REPEL-D & 14.00 & 5.34 & 7.73 & 17.00 & 12.98 & 14.72 & 6.00 & 1.35 & 2.20 & 7.00 & 3.15 & 4.34 \\
    	    REPEL-P & 64.00 & 24.43 & 35.36 & 70.00 & 53.44 & 60.61 & 32.00 & 7.21 & 11.77 & 33.00 & 14.86 & 20.49 \\ 
    	    REPEL & \textbf{78.00} & \textbf{29.77} & \textbf{43.09} & \textbf{76.00} & \textbf{58.02} & \textbf{65.80} & \textbf{48.00} & \textbf{10.81} & \textbf{17.65} & \textbf{43.00} & \textbf{19.37} & \textbf{26.71} \\ \hline
	    \end{tabular}
	}
	\end{center}
\end{table*}

\begin{figure}[htb!]
	\centering
	\subfigure[Wiki]{
		\label{fig::pr-wiki}
		\includegraphics[width=0.22\textwidth]{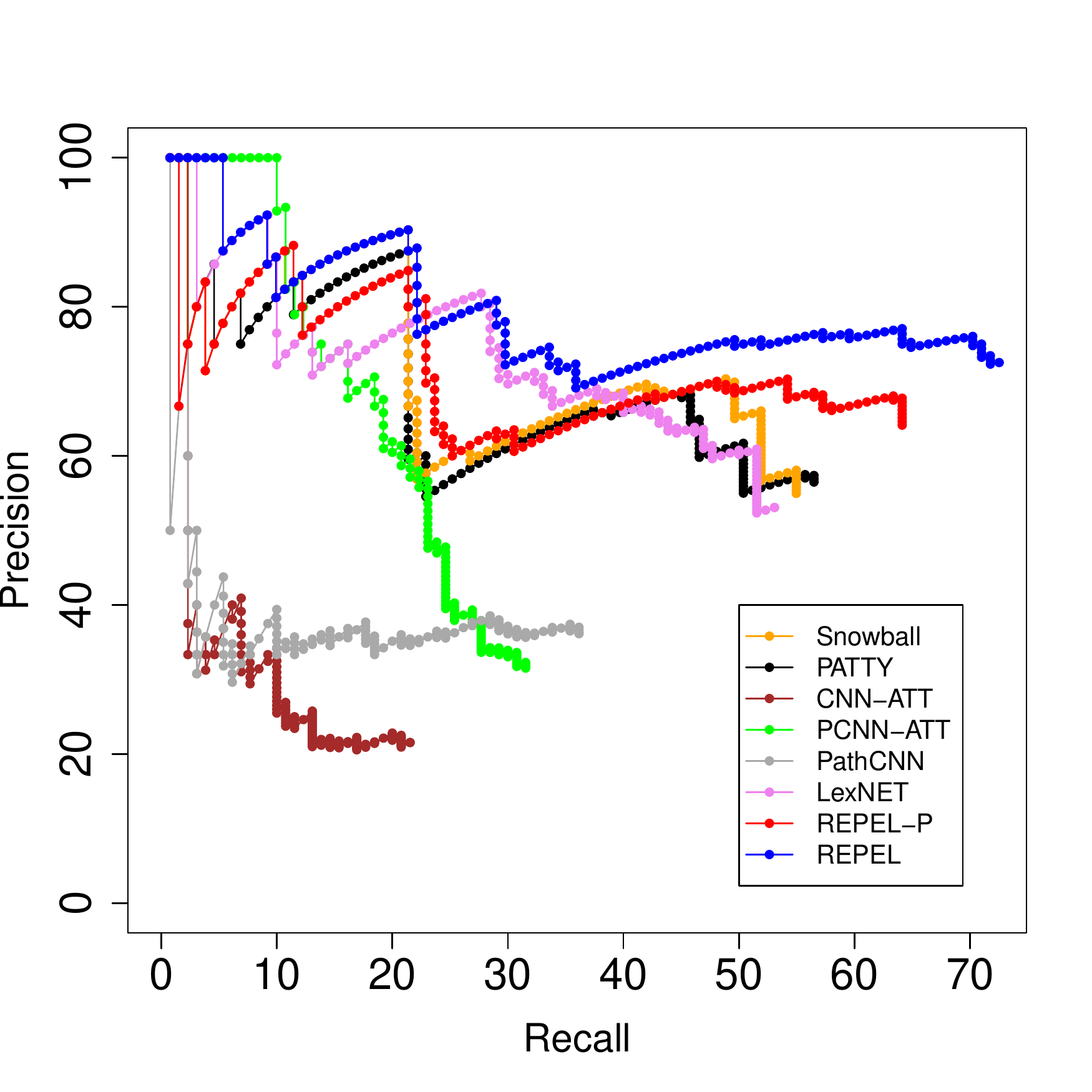}	
	}
	\subfigure[NYT]{
		\label{fig::pr-nyt}
		\includegraphics[width=0.22\textwidth]{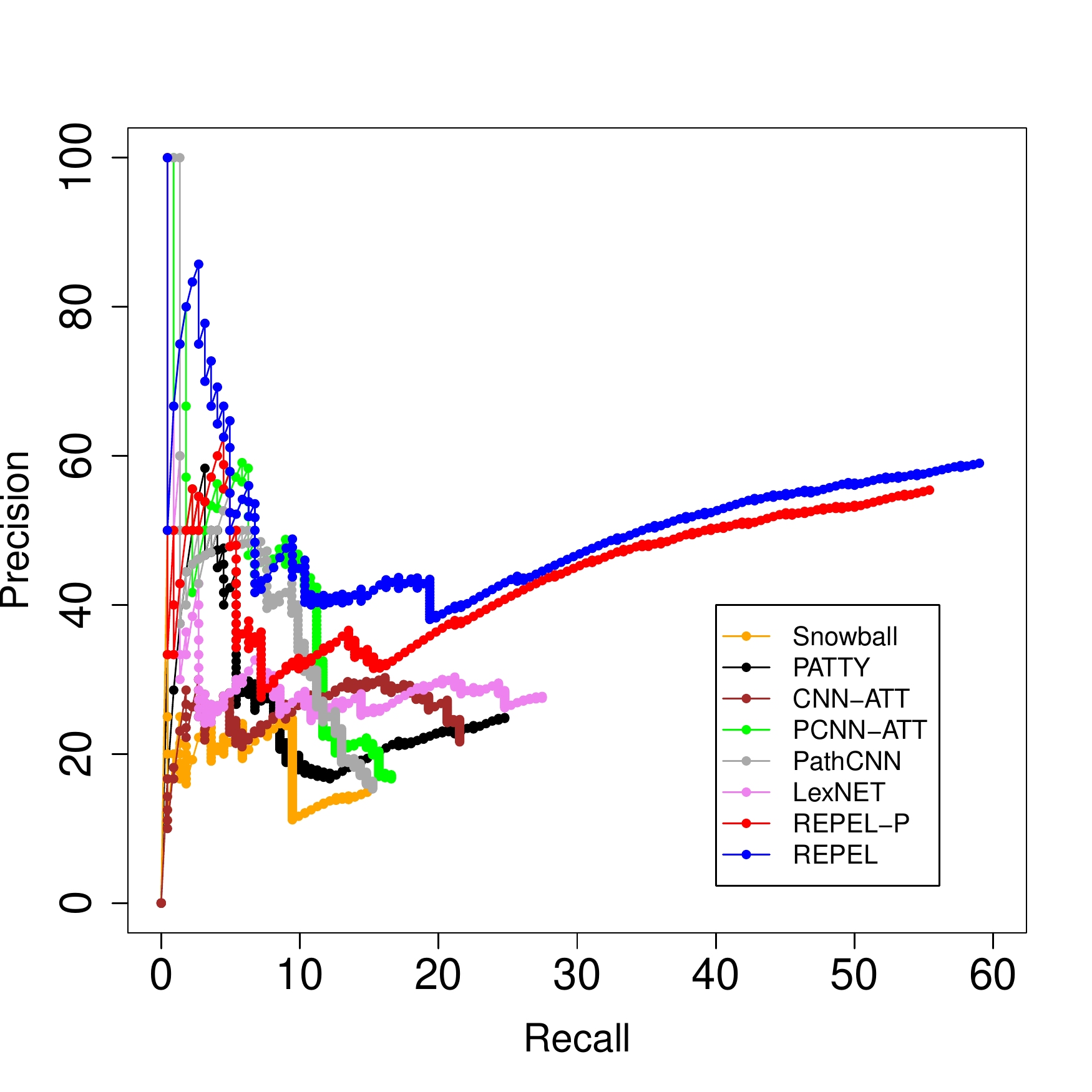}
	}
	\caption{Precision-Recall curves on the RE task.}
	\label{fig::pr}
\end{figure}

\smallskip
\noindent \textsf{\textbf{2. Corpus-level Relation Extraction (RE).}}
Next, we show the results on the corpus-level relation extraction task. We present the quantitative results in Table~\ref{tab::results-re} 
and the precision-recall in Fig.~\ref{fig::pr}. For the approaches using textual patterns (PATTY, Snowball), we see the results are quite limited especially on the NYT dataset. This is because it discovers informative patterns in a bootstrapping way, which can lead to the semantic drift problem~\cite{curran2007minimising} and thus harm the performance. For other neural network based pattern approaches (PathCNN, CNN-ATT, PCNN-ATT), although they are proved to be very effective when the given instances are abundant, their performance in the weakly-supervised setting is not satisfactory. The reason is that they typically deploy complicated convolutional layers or recurrent layers in their model, which rely on massive relation instances to tune. However, in our setting, the instances are very limited, leading to their poor performance. For the integration approach (LexNET), although it incorporates the distributional information, the performance is still quite limited especially on the NYT dataset. This is because the joint training framework of LexNET also requires considerable training instances.

For our proposed approach, the performance of the distributional module (REPEL-D) is very bad. This is because each test entity is mentioned in only few test sentences, and thus the learned entity representations are not so effective due to the sparsity of the distributional information. On the other hand, the pattern module (REPEL-P) of our approach achieves surprisingly good results, which are comparable to the neural models. This is because we represent each pattern using the average embedding of tokens in the pattern for pattern matching, where the token embedding is learned from the given text corpus. Although such strategy is very naive compared with the neural encoding methods, it does not involve any extra parameters to learn. In the weakly-supervised setting, the neural methods are usually hard to train due to the large number of parameters, leading to inferior results. Whereas our approach achieves impressive results because of its simplicity. Furthermore, comparing the pattern module (REPEL-P) with the complete framework (REPEL), we see that the complete framework further outperforms the pattern module, which demonstrates that the distributional module can also enhance the pattern module by helping estimate pattern reliability. 

Overall, in the weakly-supervised setting, our approach is able to achieve comparable results compared with the neural methods. Besides, the distributional module can indeed improve the pattern module with our co-training framework.

\subsection{Performance Analysis}

\smallskip
\noindent \textsf{\textbf{1. Performance w.r.t. the Number of Seed Instances.}}
To overcome the challenge of seed scarcity, our approach encourages both modules to provide extra supervision for each other. 
In this section, we thoroughly study whether our framework is indeed robust to the scarcity of seed instances. We take the Wiki dataset as an example, and report the performance of different methods under differ number of seed instances.

\begin{figure}[htb!]
	\centering
	\subfigure[KBC]{
		\label{fig::sparsity-kbc}
		\includegraphics[width=0.22\textwidth]{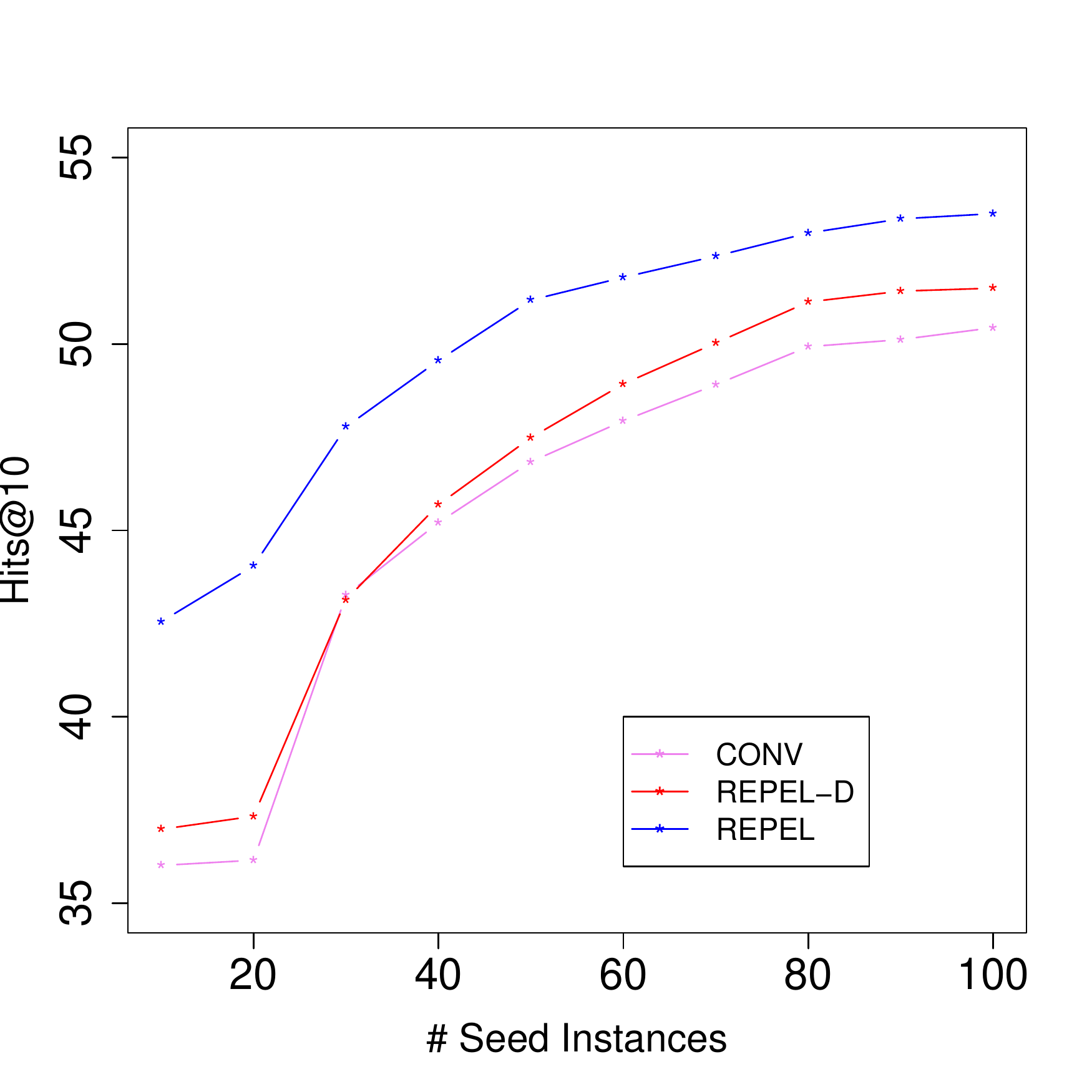}	
	}
	\subfigure[RE]{
		\label{fig::sparsity-re}
		\includegraphics[width=0.22\textwidth]{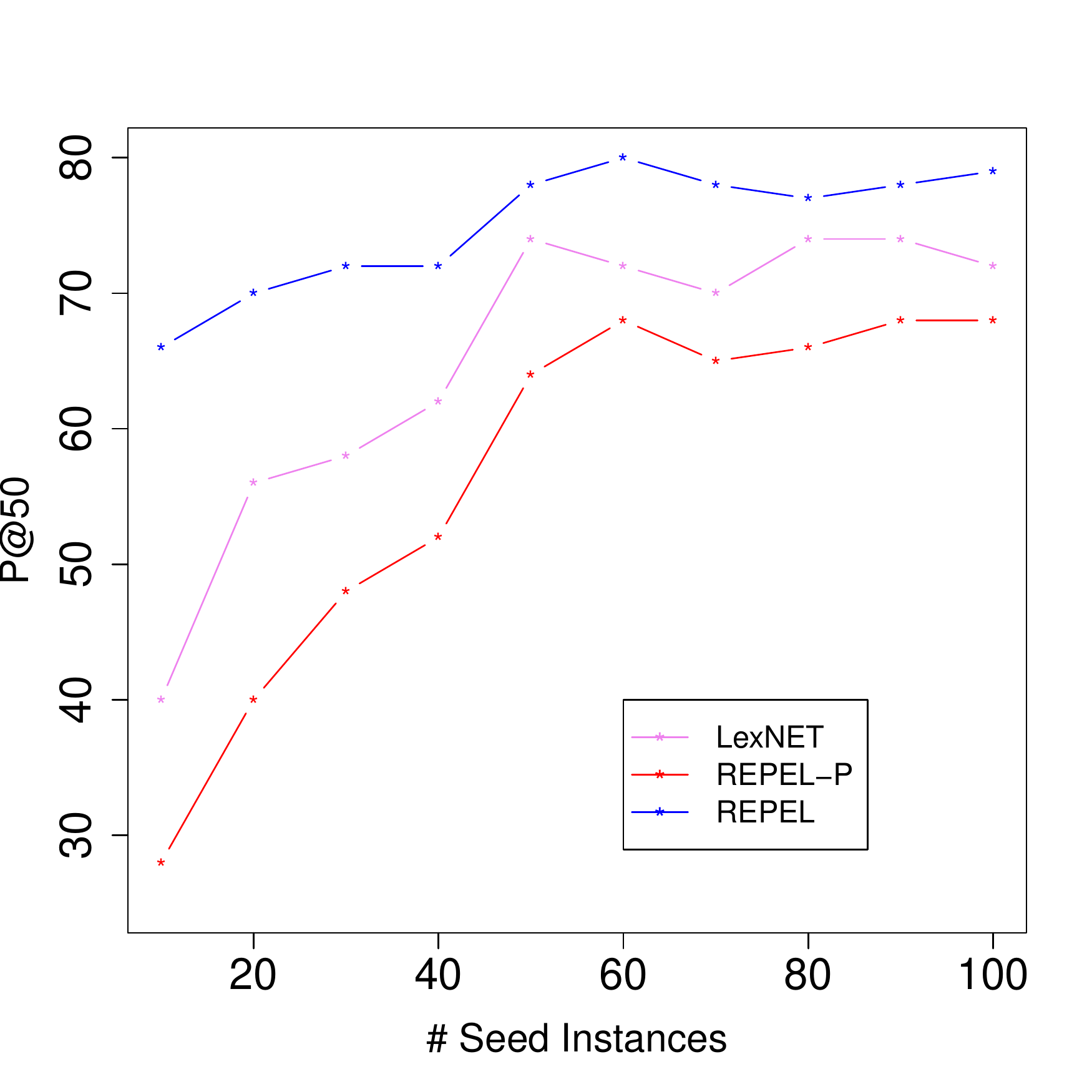}
	}
	\caption{Performance w.r.t. \# relation instances. Our approach consistently outperforms the compared algorithms especially when the given seeds are very limited.}
	\label{fig::sparsity}
\end{figure}

Fig.~\ref{fig::sparsity} presents the results on the KBC and RE tasks. We see that our approach (REPEL) consistently outperforms other approaches (CONV, LexNET) integrating both the distributional and pattern-based methods. Besides, our approach (REPEL) also achieves better results than its variants (REPEL-P, REPEL-D), which deploy only one module. Moreover, we observe that when the given seed instances are quite sufficient, the results of different approaches are pretty close. Whereas under very limited seed instances, our approach (REPEL) significantly outperforms its variants (REPEL-P, REPEL-D) and the baseline approaches (CONV, LexNET). Based on the observation, we see that with the co-training framework, our approach is more robust to seed scarcity compared with existing integration approaches (CONV, LexNET).

\smallskip
\noindent \textsf{\textbf{2. Convergences Analysis.}}
In our approach, we leverage the coordinate gradient descent algorithm for optimization, alternating between updating the distributional module and improving the pattern module. Next, we examine the optimization algorithm and study whether it converges during training. We take the Wiki dataset as an example, and present the performance of our approach at each iteration.

\begin{figure}[htb!]
	\centering
	\subfigure[KBC]{
		\label{fig::conv-wiki}
		\includegraphics[width=0.22\textwidth]{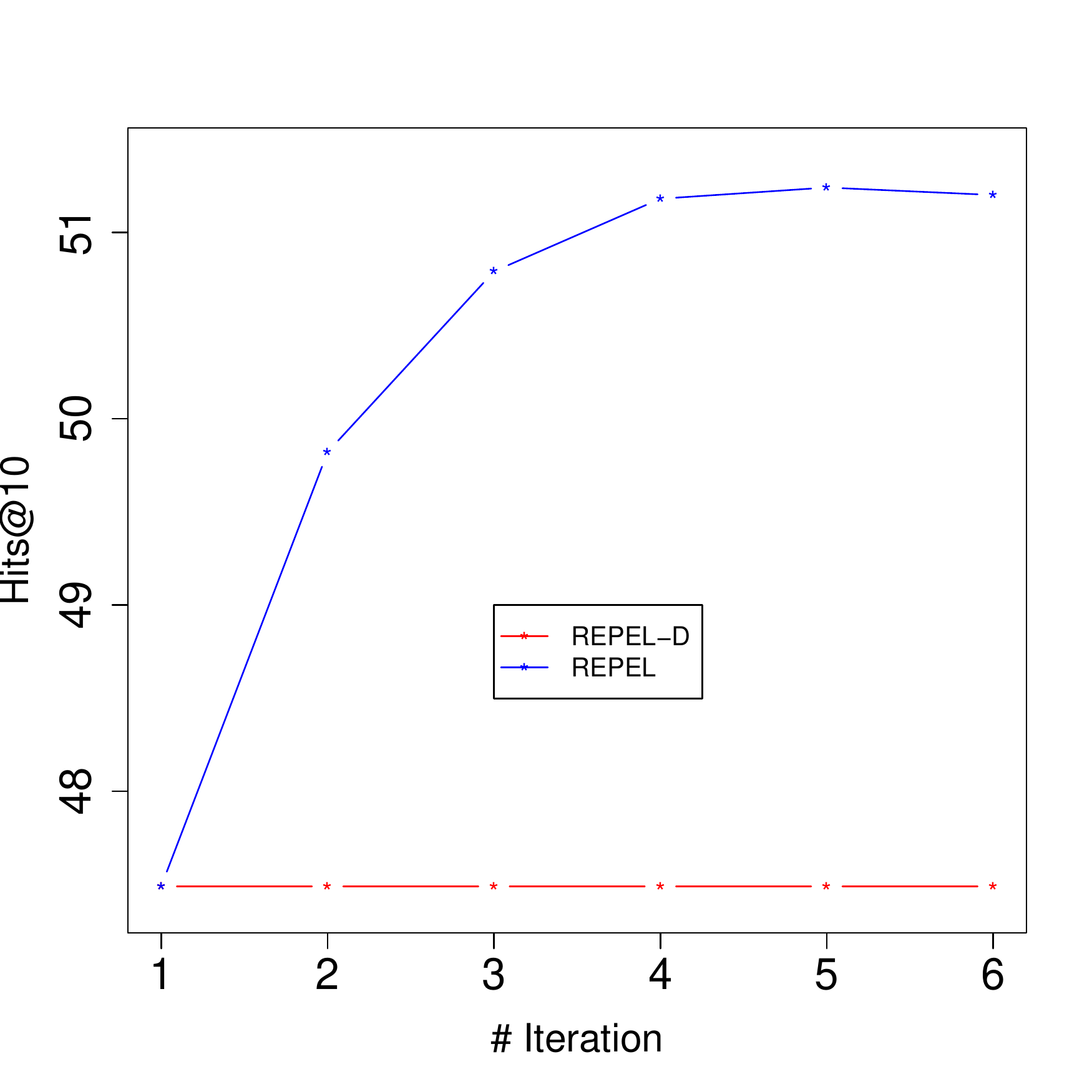}	
	}
	\subfigure[RE]{
		\label{fig::conv-nyt}
		\includegraphics[width=0.22\textwidth]{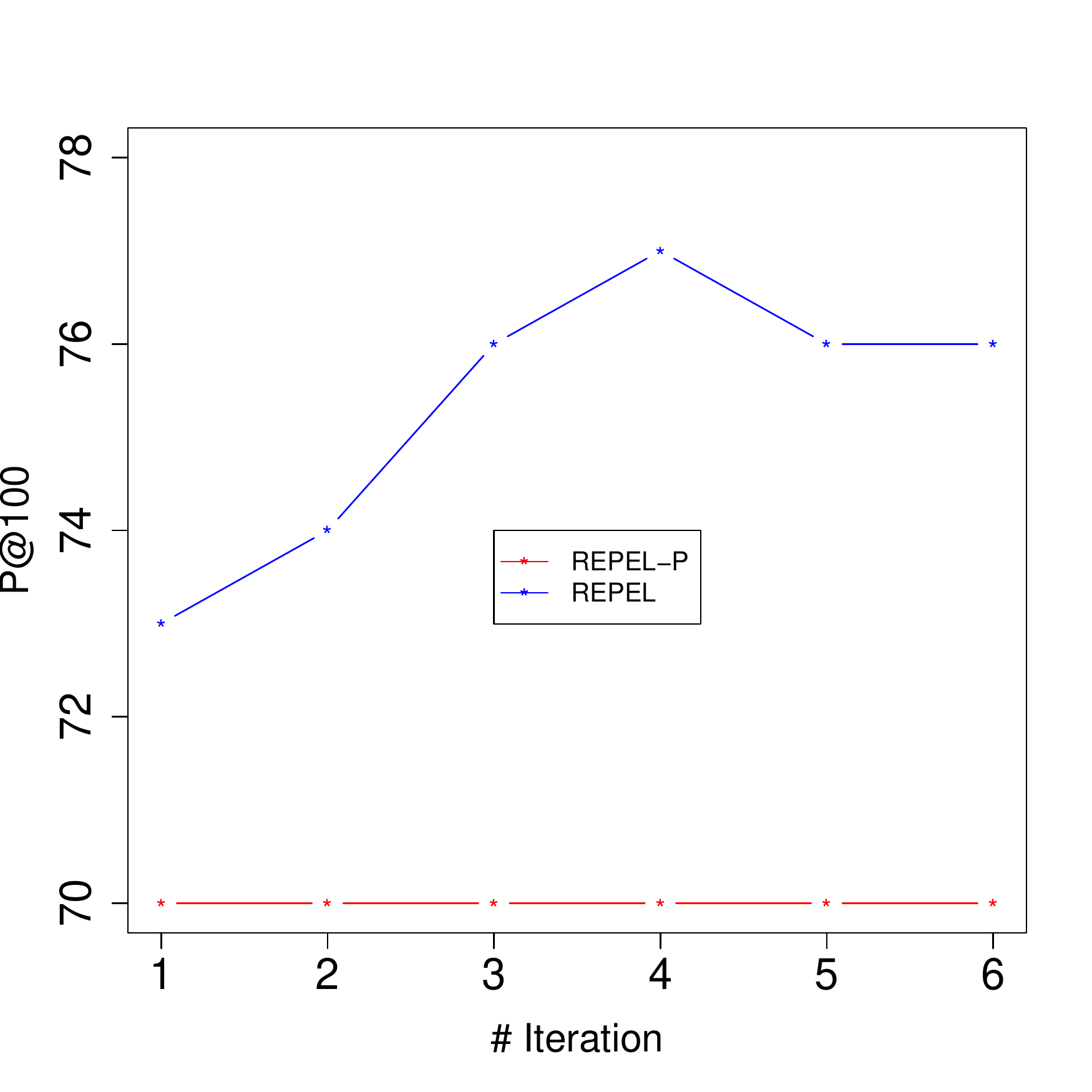}
	}
	\caption{Convergence curves of our approach. Our approach quickly converges after several iterations.}
	\label{fig::conv}
\end{figure}

Fig.~\ref{fig::conv} presents the results. In both tasks, the performance of our approach is consistently improved at the first several iterations, which shows that both modules can keep improving each other in our framework. Besides, we see that our approach quickly converges after several (3$\sim$4) iterations, which demonstrates the efficiency of the optimization algorithm.

\smallskip
\noindent \textsf{\textbf{3. Performance w.r.t. $\lambda$.}}
In our framework, the parameter $\lambda$ controls the weight of the interaction term $O_i$ (Eqn.~\ref{eqn::o_i}). A large $\lambda$ encourages strong interactions of both modules, whereas a small $\lambda$ corresponds to weak interactions. In this part, we study the performance of our approach under different $\lambda$. We take the Wiki dataset as an example, and report the results on both tasks.

\begin{figure}[htb!]
	\centering
	\subfigure[KBC]{
		\label{fig::lambda-kbc}
		\includegraphics[width=0.22\textwidth]{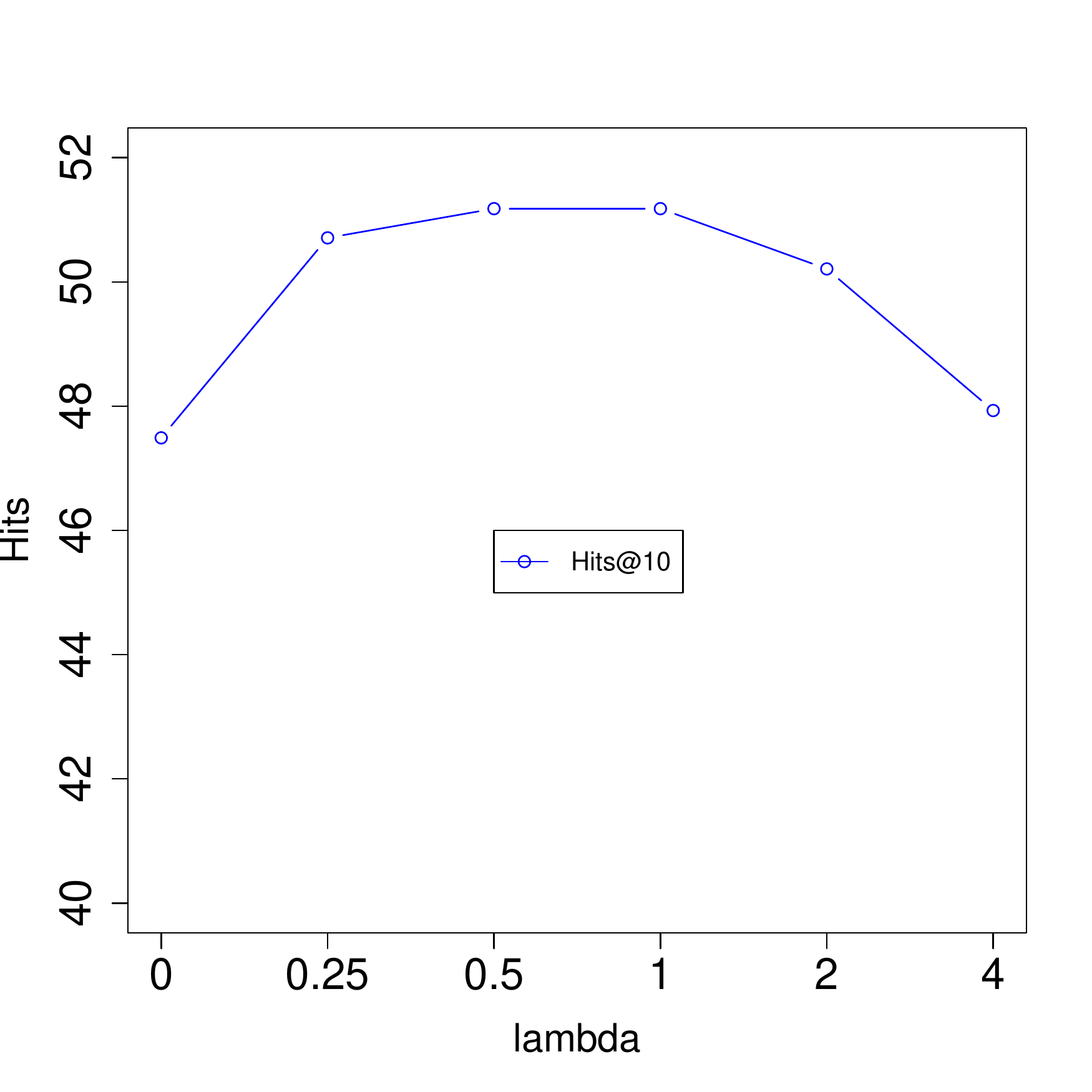}	
	}
	\subfigure[RE]{
		\label{fig::lambda-re}
		\includegraphics[width=0.22\textwidth]{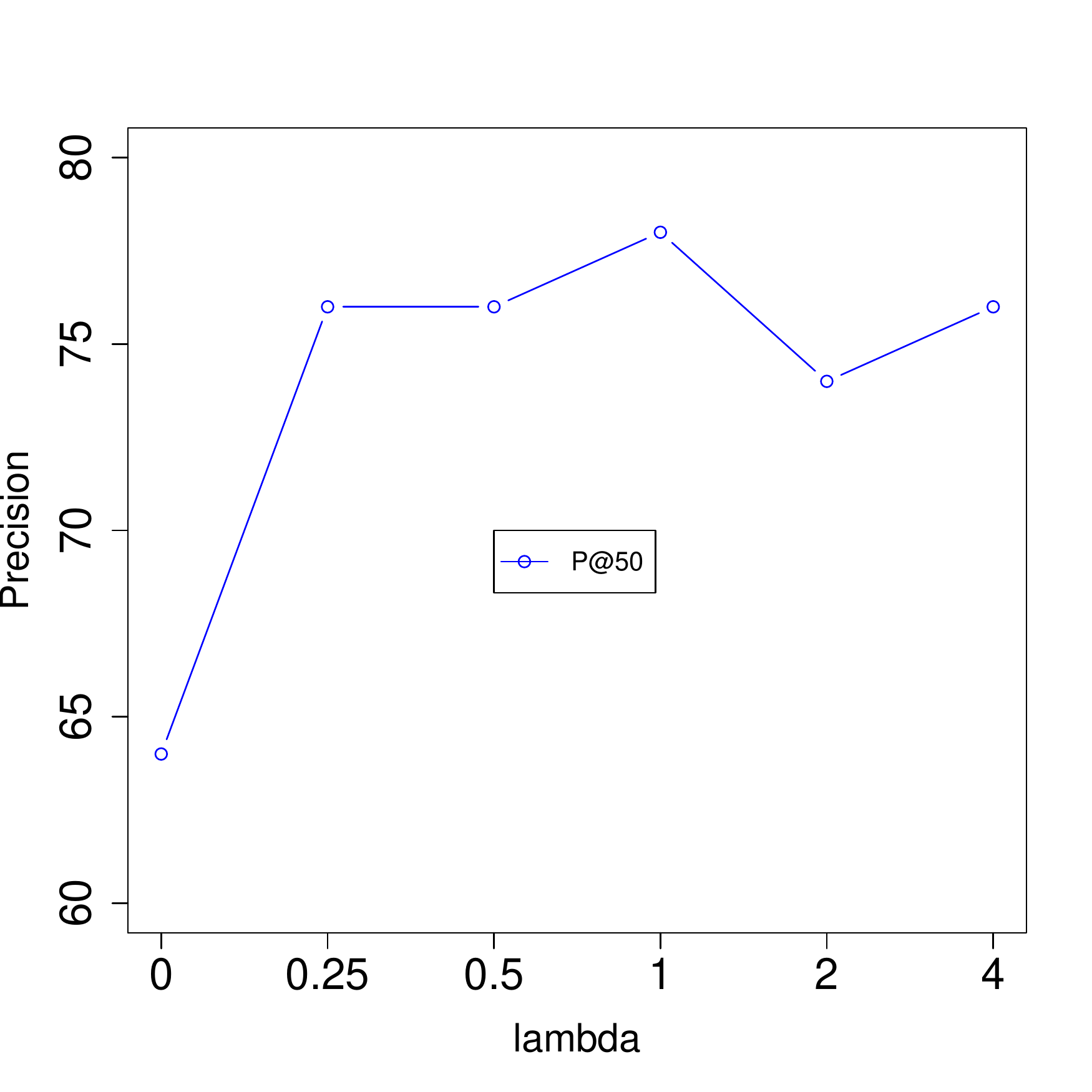}
	}
	\caption{Performances w.r.t. $\lambda$. Encouraging the interaction of the both modules ($\lambda > 0$) improves the performance.}
	\label{fig::lambda}
\end{figure}

Fig.~\ref{fig::lambda} presents the results. When $\lambda$ is set as 0, meaning that there is no interaction between the modules, we see that the results are quite limited. Then the results are quickly improved as we gradually increase $\lambda$, which further remain stable in the range (0.5, 1). If we further increase $\lambda$, the results begin to drop in the knowledge base completion task, as a large $\lambda$ emphasizes too much on the supervision provided by the modules, and thus ignores the supervision from the given seed instances.

\smallskip
\noindent \textsf{\textbf{4. Case Study.}}
In our co-training framework, both modules will collaborate with each other to overcome the seed scarcity problem. Specifically, the distributional module provides extra signals to select reliable patterns, whereas the pattern module discovers some highly confident instances to improve the distributional module. Next, we show some case study results to intuitively illustrate that both modules can indeed mutually enhance each other.

\begin{table}[!htb]
\caption{The most reliable patterns discovered by our approach. Blue patterns are incorrect ones by human.}
	\label{tab::pattern}
	\centering
	\scalebox{0.9}{
        \begin{tabular}{|C{4.35cm}|C{4.35cm}|}
        \hline
        \multicolumn{2}{|c|}{\textbf{Relation: people.person.place-of-birth}} \\
        \hline
        \textbf{REPEL-P} & \textbf{REPEL}  \\
        \hline
        $[$Tail$]$ $[$Head$]$ birthplace & $[$Head$]$ born place $[$Tail$]$ \\
        $[$Head$]$ born place $[$Tail$]$ & $[$Head$]$ born city $[$Tail$]$ \\
        \color{blue}{$[$Head$]$ father move $[$Tail$]$} & $[$Tail$]$ $[$Head$]$ birthplace \\
        $[$Head$]$ born city $[$Tail$]$ & $[$Head$]$ born June $[$Tail$]$ \\
        $[$Head$]$ born January $[$Tail$]$ & $[$Head$]$ live return $[$Tail$]$ \\
        \hline
        \hline
        \multicolumn{2}{|c|}{\textbf{Relation: people.person.parents}} \\
        \hline
        \textbf{REPEL-P} & \textbf{REPEL}  \\
        \hline
        $[$Tail$]$ die succeed son $[$Head$]$ &  $[$Tail$]$ die succeed son $[$Head$]$ \\
        \color{blue}{Babur $[$Tail$]$ $[$Head$]$} &  $[$Head$]$ daughter $[$Tail$]$ \\
        $[$Tail$]$ give boy $[$Head$]$ &  $[$Head$]$ son $[$Tail$]$ \\
        $[$Head$]$ son $[$Tail$]$ &  descendant $[$Tail$]$ $[$Head$]$ \\
        \color{blue}{have relationship $[$Head$]$ $[$Tail$]$} &  $[$Tail$]$ marry have son $[$Head$]$ \\
        \hline
        \end{tabular}
    }
     \vspace{-0.3cm}
\end{table}

We first present the most reliable path-based patterns (i.e., tokens along the shortest dependency path between two entities) discovered by our approach and its variant on the Wiki dataset in Table~\ref{tab::pattern}. Blue patterns are unreliable ones based on the human. Comparing our approach with its variant (REPEL-P), we see that by considering the supervision signals from the distributional module (REPEL), some unreliable patterns can be filtered out from the pattern list, and the patterns discovered by our approach (REPEL) are more reliable. Therefore, the distributional module can indeed help the pattern module for reliable pattern selection.

\begin{table}[!htb]
\caption{Top ranked instances extracted by the reliable patterns. Blue instances are incorrect ones by human.}
	\label{tab::fact}
	\centering
	\scalebox{0.9}{
        \begin{tabular}{|C{8.7cm}|}
        
        \hline
        \textbf{Relation: people.person.nationality} \\
        \hline
        (Charles IV of France, France) (Benjamin Franklin, USA) \\
        (Adolf Hitler, German)  (Thomas Jefferson, USA) \color{blue}{(Pol Pot, Thailand)} \\
        \hline
        
        
        
        
        \hline
        \textbf{Relation: location.country.capital} \\
        \hline
        (Denmark, Copenhagen) \color{blue}{(Vietnam, Ho Chi Minh City)} \\
        (Norway, Oslo) (Yugoslavia, Belgrade) (Guinea, Conakry) \\
        \hline
        
        \end{tabular}
    }
    \vspace{-0.3cm}
\end{table}

Meanwhile, we also randomly sample some instances extracted by the discovered reliable patterns, and we show them in Table~\ref{tab::fact}, where the blue instances are the incorrect ones by human. From the results, we see that most instances extracted by the reliable patterns are correct and reasonable. Therefore, the pattern module can in turn benefit the distributional module by providing some reasonable relation instances.

\section{RELATED WORK}
\label{sec::related}

Our work is related to pattern-based approaches for relation extraction. Given two entities, the pattern-based approaches predict their relation from sentences mentioning both entities. Traditional approaches~\cite{agichtein2000snowball,nakashole2012patty,schmitz2012open,yahya2014renoun,jiang2017metapad} try to find some informative textual patterns using the given instances, and utilize the patterns for extraction. However, these approaches ignore the semantic correlations of patterns, and thus suffer from semantic drift~\cite{curran2007minimising}. Recent approaches~\cite{xu2015classifying,liu2015dependency,toutanova2015representing,lin2016neural,shwartz2016improving,zeng2017incorporating} address the problem by encoding textual patterns with neural networks. Despite their success, these approaches rely on considerable labeled instances to train effective models, which suffer from the seed scarcity problem in the weakly-supervised setting. Our approach solves the problem by letting the distributional module provide extra supervision.

Our work is also related to the distributional approaches. Typically, these approaches learn entity representations from corpus-level statistics, and meanwhile a relation classifier is trained with the relation instances, which takes entity representations as features for relation prediction. Some approaches learn entity representations from only text corpora~\cite{mikolov2013distributed,pennington2014glove,tang2015line}. However, their performances are usually limited due to the lack of supervision. Some other approaches~\cite{bordes2013translating,yang2014embedding,ji2015knowledge,lin2015learning,xu2014rc,wang2014knowledge,wang2016solving} learn more predictive entity representations by using the given relation instances as supervision, achieving superior results. However, they also require abundant relation instances to learn effective relation classifiers, which are hard to obtain in the weakly-supervised setting. Our approach alleviates the problem by letting the pattern module to generate some highly confident instances as extra seeds.

There are also handful studies~\cite{riedel2013relation,toutanova2015representing,shwartz2016improving,verga2016generalizing,qu2017automatic} trying to integrate the distributional and pattern-based approaches. Typically, they jointly train a distributional model and a pattern model. However, the supervision of each model totally comes from the given relation instances, which is insufficient in the weakly-supervised setting. Our approach solves the seed scarcity problem with a co-training framework, which encourages both models to provide extra supervision for each other.
\section{CONCLUSIONS}

In this paper, we studied corpus-level relation extraction in the weakly-supervised setting. We proposed a novel co-training framework called REPEL to integrate a pattern module and a distributional module. Our framework encouraged both modules to provide extra supervision for each other, so that they can collaborate to overcome the scarcity of seeds. Experimental results proved the effectiveness of our framework. In the future, we plan to enhance the pattern module by using neural models for pattern encoding.


\bibliographystyle{abbrv}
\bibliography{sigproc}
\end{document}